\documentclass[sn-apa,iicol]{sn-jnl}% APA Reference Style 
\usepackage{graphicx}%
\usepackage{amsmath,amssymb,amsfonts}%
\usepackage{amsthm}%
\usepackage{mathrsfs}%
\usepackage[title]{appendix}%
\usepackage{xcolor}%
\usepackage{textcomp}%
\usepackage{manyfoot}%
\usepackage{booktabs}%
\usepackage{algorithm}%
\usepackage{algorithmicx}%
\usepackage{algpseudocode}%
\usepackage{listings}%
\usepackage{subfigure}
\usepackage{times}
\usepackage{utfsym}
\usepackage{diagbox}
\usepackage{bbding}
\usepackage{makecell}
\usepackage{bm}
\usepackage{url}
\usepackage{color}
\usepackage{epsfig,epstopdf}
\usepackage{pifont}
\usepackage{multirow}
\usepackage{tabularx,arydshln}
\usepackage[accsupp]{axessibility}  % Improves PDF readability for those with disabilities.
\theoremstyle{thmstyleone}%
\theoremstyle{thmstyletwo}%

\theoremstyle{thmstylethree}%

\begin{document}

\title[Article Title]{PosMLP-Video: Spatial and Temporal Relative Position Encoding for Efficient Video Recognition}

%%=============================================================%%
%% Prefix	-> \pfx{Dr}
%% GivenName	-> \fnm{Joergen W.}
%% Particle	-> \spfx{van der} -> surname prefix
%% FamilyName	-> \sur{Ploeg}
%% Suffix	-> \sfx{IV}
%% NatureName	-> \tanm{Poet Laureate} -> Title after name
%% Degrees	-> \dgr{MSc, PhD}
%% \author*[1,2]{\pfx{Dr} \fnm{Joergen W.} \spfx{van der} \sur{Ploeg} \sfx{IV} \tanm{Poet Laureate} 
%%                 \dgr{MSc, PhD}}\email{iauthor@gmail.com}
%%=============================================================%%

% \author*[1,2]{\fnm{Yanbin} \sur{Hao}}\email{haoyanbin@hotmail.com}
\author[1]{\fnm{Yanbin} \sur{Hao}}\email{haoyanbin@hotmail.com}
\equalcont{These authors contributed equally to this work.}
\author[1]{\fnm{Diansong} \sur{Zhou}}\email{zhouds1918@gmail.com}
\equalcont{These authors contributed equally to this work.}

\author[1]{\fnm{Zhicai} \sur{Wang}}\email{wangzhic@mail.ustc.edu.cn}

\author[2]{\fnm{Chong-Wah} \sur{Ngo}}\email{cwngo@smu.edu.sg}

% \author[1]{\fnm{Xiangnan} \sur{He}}\email{xiangnanhe@gmail.com}

\author[3]{\fnm{Meng} \sur{Wang}}\email{eric.mengwang@gmail.com}

\affil[1]{\orgdiv{School of Information Science and Technology, School of Artificial Intelligence and Data Science}, \orgname{University of Science and Technology of China}, \orgaddress{\street{No.96, JinZhai Road}, \city{Hefei}, \postcode{230026}, \state{Anhui}, \country{China}}}

\affil[2]{\orgdiv{School of Computing and Information Systems}, \orgname{Singapore Management University}, \orgaddress{\street{80 Stamford Road}, \postcode{178902}, \country{Singapore}}}

\affil[3]{\orgdiv{School of Computer Science and Information Engineering}, \orgname{Hefei University of Technology}, \orgaddress{\street{No. 485, Danxia Road}, \city{Hefei }, \postcode{230601}, \state{Anhui}, \country{China}}}

%%==================================%%
%% sample for unstructured abstract %%
%%==================================%%

\abstract{In recent years, vision Transformers and MLPs have demonstrated remarkable performance in image understanding tasks. However, their inherently dense computational operators, such as self-attention and token-mixing layers, pose significant challenges when applied to spatio-temporal video data. To address this gap, we propose PosMLP-Video, a lightweight yet powerful MLP-like backbone for video recognition. Instead of dense operators, we use efficient relative positional encoding (RPE) to build pairwise token relations, leveraging small-sized parameterized relative position biases to obtain each relation score. Specifically, to enable spatio-temporal modeling, we extend the image PosMLP's positional gating unit to temporal, spatial, and spatio-temporal variants, namely PoTGU, PoSGU, and PoSTGU, respectively. These gating units can be feasibly combined into three types of spatio-temporal factorized positional MLP blocks, which not only decrease model complexity but also maintain good performance. Additionally, we enrich relative positional relationships by using channel grouping. Experimental results on three video-related tasks demonstrate that PosMLP-Video achieves competitive speed-accuracy trade-offs compared to the previous state-of-the-art models. In particular, PosMLP-Video pre-trained on ImageNet1K achieves 59.0\%/70.3\% top-1 accuracy on Something-Something V1/V2 and 82.1\% top-1 accuracy on Kinetics-400 while requiring much fewer parameters and FLOPs than other models. The code is released at \url{https://github.com/zhouds1918/PosMLP_Video}.}


%%\pacs[JEL Classification]{D8, H51}

%%\pacs[MSC Classification]{35A01, 65L10, 65L12, 65L20, 65L70}

\maketitle

\section{Introduction}

Video neural networks are typically built upon image neural networks. One common approach for processing 3D video signals is to extend 2D spatial operators to a 3D version by including the time axis. Examples of such networks include C3D \cite{tran2015learning} and I3D \cite{carreira2017quo} networks. However, this straightforward extension leads to a significant increase in model complexity and processing time due to the addition of the time axis. To alleviate this problem, video architectures adopt the factorization of space and time. For instance, P3D \cite{qiu2017learning} decomposes the 3D convolution kernel as the combination of 1D kernel + 2D kernel. Similarly, ViViT \cite{arnab2021vivit}, MorphMLP \cite{zhang2022morphmlp} and MLP-3D \cite{qiu2022mlp} divide the 3D spatio-temporal tokens as spatial tokens and temporal tokens. In addition to the factorization mechanism, there are also other strategies, such as temporal \cite{lin2019tsm} or token shift \cite{zhang2021token}, that aim to create lightweight models. Despite these well-designed mechanisms, video models still face challenges due to the large number of parameters and heavy computational burden arising from dense spatial-temporal computational operators like self-attention and token-mixing layers. These dense operators contribute to the overall model complexity and computational requirements.

This paper presents a novel idea of leveraging relative position encoding (RPE) to compute spatial and temporal relations for achieving a linear model complexity. The proposed lightweight video model architecture, called PosMLP-Video, is highly efficient for requiring only $2N-1$ relation parameters for a given $N$ tokens compared to other models~\cite{zhang2022morphmlp,qiu2022mlp}. Different from the token-mixing in vision MLPs and the self-attention in Transformers, the computation of pairwise token relations is reduced by searching from a learnable relative position bias dictionary with minimal parameters. While a single dictionary for a set of tokens determines the relation score of any two tokens based solely on their relative position difference, introducing multiple dictionaries can help maintain diverse positional relationships and enhance the diversity of positional inductive biases. We thus split tokens into multiple groups along the channel dimension and learn a specific relative position bias dictionary for each group, resulting in a total of $g$ (number of groups) dictionaries.  PosMLP-Video builds upon the image PosMLP \cite{wang2022parameterization} and extends the static spatial cross-token relation computation to handle videos. With PosMLP-Video, we explore the potential of learnable relative position encoding (LRPE) in constructing spatial and temporal token relations, which is more suitable for videos than  GQPE used in the image PosMLP. Moreover, we investigate various space-time factorized designs to obtain optimal complexity-accuracy trade-offs.

PostMLP-Video is implemented as a hierarchical MLP-link architecture for the consideration of the high efficiency in both training and inference~\cite{qiu2022mlp}. It begins with a patch embedding layer for processing the input video frames. It then incorporates four stages of video PosMLP blocks, with a spatial downsampling layer between each stage. The architecture of video PosMLP blocks considers three \textbf{design questions (DQ)} involving space-to-time extension, spatio-temporal modeling, and task generalization, as followings. 

% As such, this channel grouping operation increases the LRPE-based model complexity to $g$ (number of groups) times, i.e., $O(gN)$. This is different to the group operation in convolution which will reduce the parameters by $g$.relationships, 

\begin{itemize}
  \item \textbf{DQ1:} \textbf{\textit{Can the spatial operator of PosMLP be directly applied to process the temporal axis?}} This question may not pose a significant challenge for CNNs and Transformers as convolutional operations focus on local areas, while self-attention captures global contexts. However, as demonstrated by the PosMLP \cite{wang2022parameterization}, the computation unit based on LRPE exhibits a strong locality while still possessing a global receptive field. We can also extend this question to ask \textbf{\textit{whether we should control the receptive field of PosMLP on the temporal axis.}}

  In implementation, we adapt the original 2D positional spatial gating unit (\textbf{PoSGU}) of PosMLP to a 1D positional temporal gating unit (\textbf{PoTGU}) and a 3D positional spatio-temporal gating unit (\textbf{PoSTGU}). These units leverage LRPE to compute pairwise token relations and incorporate a part-to-part element-wise gating mechanism as gMLP \cite{liu2021pay} for cross-token interactions. Each of the three units captures axis-preferred cross-token interactions. Moreover, we use the window partitioning \cite{liu2021swin} strategy to control the spatial and temporal receptive fields to evaluate how the performance changes. Finally, experimental results demonstrate that prorate window partitioning for spatial operation can result in superior efficiency-accuracy trade-off while a small window in the time axis will significantly decrease performance.

  \item \textbf{DQ2:} \textbf{\textit{Are the existing 2D-to-3D mechanisms in video CNNs and Transformers suitable for PosMLP-Video, and how can we implement these mechanisms?}} Exploring this question involves considering the implementation of the axis-extended 3D spatio-temporal computational operation and the lightweight spatio-temporal factorization strategies for PosMLP-Video. By addressing this question, we can demonstrate the potential for the LRPE-based computation unit to become a general operator, akin to convolution or self-attention.

  We present three spatio-temporal factorized video PosMLP blocks: two cascaded blocks (\textbf{PoTGU$\rightarrow$PoSGU} and \textbf{PoSGU$\rightarrow$PoTGU}), one paralleled block (\textbf{PoTGU+PoSGU}), and the joint spatio-temporal video PosMLP block (\textbf{PoSTGU}) for comparison. Experimental results demonstrate that the spatio-temporal factorized variants have smaller model sizes and achieve higher recognition accuracies compared to the joint version. Among these variants, the paralleled version \textbf{PoTGU+PoSGU} achieves the best performance. Moreover, we also investigate various kinds of combinations for the spatial and temporal features of \textbf{PoTGU+PoSGU} to find a more suitable design.

  \item \textbf{DQ3:} \textbf{\textit{Can the proposed PosMLP-Video be utilized for tasks beyond general action recognition?}} PosMLP-Video is proposed for video content modeling. Apart from recognizing general actions or activities, we investigate its potential capability of locating actions and recognizing facial micro-expressions. This can further enhance its versatility and demonstrate its broader utility. In the experiment, we additionally conduct experiments on action detection and micro-expression recognition tasks. The action detection requires the model to locate and identify multiple spatio-temporal actions within a given video. Recognizing micro-expressions requires the model to be sensitive to subtle facial dynamics.
\end{itemize}

Through the presentation of three insightful design questions, we attain a thorough comprehension of the adaptation of the image PosMLP to the video domain. As a novel MLP-like video neural network backbone, our exploration encompasses the design philosophy of the new fundamental LRPE-based spatio-temporal operators and the corresponding stage-level spatio-temporal blocks, and the application across multiple video understanding tasks. Our PosMLP-Video variants provide  new insights for MLP-like video backbones.

% \item \textbf{DQ3:} \textbf{\textit{Is it beneficial to create multiple relative position bias dictionaries?}} While a single dictionary for a set of tokens determines the relation score of any two tokens based solely on their relative position difference, introducing multiple dictionaries can help maintain diverse positional relationships and enhance the diversity of positional inductive biases. This allows the model to better understand and leverage various contexts.

% We adopt the channel grouping operation to achieve this goal. Specifically, we split tokens into multiple groups along the channel dimension and learn a specific relative position bias dictionary for each group. As such, this channel grouping operation increases the LRPE-based model complexity to $g$ (number of groups) times, i.e., $O(gN)$. This is different to the group operation in convolution which will reduce the parameters by $g$.

PosMLP-Video can be readily pre-trained on off-the-shelf image datasets by replacing the temporal unit PoTGU with a residual connection. In our experiment, we demonstrate that, despite being pre-trained on the relatively smaller dataset ImagNet1K, our PosMLP-Video can achieve comparable or even superior performance to competing methods that are pre-trained on more extensive datasets such as ImageNet21K and Kinetics-400/600. Overall, through various video understanding tasks, PosMLP-Video proves to be competitive in performance with better model efficiency than the existing general video models, such as CNNs, Transformers, and MLPs, and other task-specific models.

\section{Related Works}
Our PosMLP-Video relies on relative position encoding for spatial and temporal encoding. Thus, we first present a review of position encoding in sequential data modeling. Then, we introduce video neural networks that adopt spatio-temporal factorization to achieve efficient modeling. Finally, we compare our PosMLP-Video with other existing MLP-based video networks. 
% We start by introducing video neural network models, including video CNNs, Transformers and MLPs. Then a review of position encoding in sequential data modeling is presented.

{\bf Position encoding in sequential data modeling.} The order information is important for understanding sequential data such as natural language and video. However, since self-attention is typically order-independent, Transformers commonly utilize the position encoding (PE) to maintain order information \cite{raffel2020exploring}. Two widely used PE methods are absolute position encoding (APE), e.g., the sinusoidal position signal \cite{vaswani2017attention} and learned position embeddings, and relative position encoding (RPE), e.g., learnable RPE (LRPE) \cite{shaw2018self} and quadratic PE (QPE) \cite{cordonnier2019relationship}. APE is usually used as a simple way of adding the signal or embedding to the corresponding token. Many text and vision Transformers \cite{arnab2021vivit,dosovitskiy2020image,vaswani2017attention,wang2022deformable,yang2022recurring,chen2022mm,yan2022multiview,fan2021multiscale,xing2023svformer,tu2023implicit} have demonstrated its positive effect on sequential modeling. In contrast to APE, RPE computes the relative position embeddings based on the relative position offset. Since the position offset exactly overlaps the key-query offset of self-attention, it is a common method for collapsing the embedding to a learnable scalar and adding it to the corresponding attention score, which is also the principle of LRPE. Compared with APE, RPE regularly provides more significant performance improvement for Transformers \cite{shaw2018self,d2021convit,liu2021swin,wang2022parameterization,liu2022video,li2022mvitv2,wu2022memvit} in language and vision tasks. In addition, despite the lack of explicit use of PE in CNNs and MLPs, it has been demonstrated that both can capture position information implicitly in their model learning \cite{islam2021global,kayhan2020translation,islam2021position,wang2022parameterization}. These findings motivate us to develop a pure RPE-based operator for efficiently processing time-order-dependent video data.

{\bf Efficient video networks through spatio-temporal factorization.} Spatio-temporal factorization is a technique that enables the separate modeling of spatial and temporal relations in video networks. By using different lightweight operators for each aspect, it offers significant reductions in both model size and computational cost compared to using a 3D operator. Representative works in this area include P3D~\cite{qiu2017learning}, R(2+1)D~\cite{tran2018closer}, S3D~\cite{xie2018rethinking}, ViViT \cite{arnab2021vivit} and TimeSformer \cite{bertasius2021space}. In P3D and S3D, two video CNNs, the 3D spatio-temporal convolution is approximated by decomposing it into a 2D spatial convolution and a 1D temporal convolution. Similarly, ViViT and TimeSformer, which are Transformer-based models, divide video attention into temporal attention and spatial attention, with each applied separately within their respective blocks. Our PosMLP-Video also employs spatio-temporal factorization for a lightweight design. However, since this strategy is being attempted for the first time in an LRPE-based video MLP network, it is essential to explore and determine a suitable design to achieve effective spatio-temporal factorization.

{\bf Video MLPs.} In recent years, vision MLPs \cite{tolstikhin2021mlp,yu2022s2,lian2021mlp,hou2022vision} merely utilize token-mixing and channel-mixing MLP to achieve cross-token and cross-channel interaction, respectively. Particularly, the token-mixing (i.e., token-FC) layer is expected to replace the self-attention layer in a Transformer. When processing video data, however, a huge number of MLP layers might still result in dense parameters and computations. The existing video MLPs include MorphMLP \cite{zhang2022morphmlp} and MLP-3D \cite{qiu2022mlp}. MorphMLP designs two types of MLP layers, i.e., MorphFC$_s$ and MorphFC$_t$, where MorphFC$_s$ captures the spatial semantics by gradually increasing the token receptive field while MorphFC$_t$ achieves long-range temporal dependency by applying a linear projection to the temporal concatenated feature chunk. The space-time factorization can significantly reduce the computation cost. MLP-3D decomposes the token mixing MLP along height, width and time axes and aggregates their outputs using weighted summation. To further improve model efficiency, they suggest a grouped time mixing operation that can also achieve parameter sharing across projections. In the approach section, we provide a comprehensive comparison between our PosMLP-Video model and the above two video MLP architectures, MorphMLP and MLP-3D.

\section{Approach}
In this section, we first introduce LRPE. Then, we elaborate on the design details of the LRPE-based spatial and temporal gating units, and the four types of video PosMLP blocks. Finally, we stack the video PosMLP blocks hierarchically to construct a set of PosMLP-Video network backbones.

\subsection{Learnable Relative Position Encoding}
RPE computes the pairwise relationships between tokens based on the position offset. Following works \cite{he2017mask,raffel2020exploring,liu2021swin,liu2022video}, a relatively efficient learnable RPE (LRPE) is introduced to determine the relevance score of each position pair by using a learned relative position bias dictionary. The relative position offset of a token pair totally dictates the positional bias ``choosing''. Therefore, LRPE is order-sensitive. We verify the sensitivity in the experiment. LRPE is often utilized to enhance self-attention. Given a 2D $M\times M$ image window as an example, the enhanced self-attention in \cite{liu2021swin} is calculated for each head as follows
\begin{equation}
    \text{Attention}(\mathbf{Q}, \mathbf{K},\mathbf{V})=\text{Softmax}\left (\frac{\mathbf{Q}\mathbf{K}^{T}}{\sqrt{d}} +\mathbf{R}\right )\mathbf{V},
\end{equation}
where $\mathbf{Q}, \mathbf{K},\mathbf{V} \in \mathbb{R}^{M^2\times d}$ denote the query, key and value matrices in a transformer, $d$ is the feature dimension, $M^2$ is the total number of tokens in a 2D window, and $\mathbf{R}\in \mathbb{R}^{M^2\times M^2}$ is the relative position bias matrix. Specifically, each element $r_{ij}$ in $\mathbf{R}$ is searched from the learned bias dictionary $\mathbf{P} \in \mathbb{R}^{(2M-1)\times (2M-1)}$ indexed by the position offset. 

\begin{figure*}[]
\centering
\includegraphics[width=0.95\textwidth]{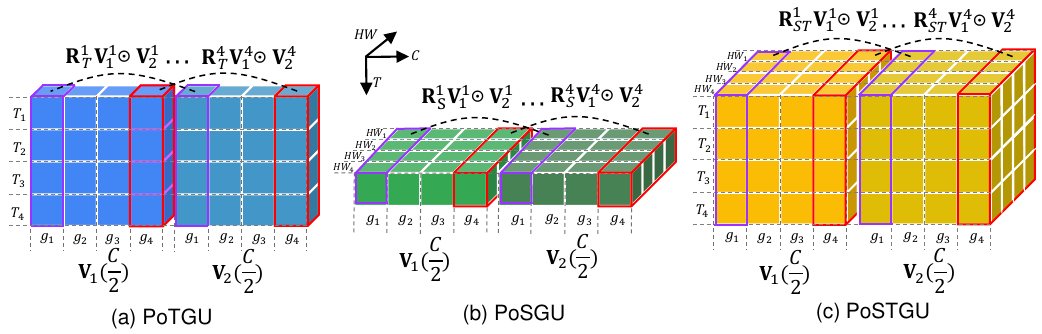}
% \vspace{-0.2cm}
\caption{Positional spatial and temporal gating units. }
\label{model_posv}
% \vspace{-0.3cm}
\end{figure*}

\subsection{Positional Spatial and Temporal Gating Units}
Utilizing the aforementioned LRPE approach, we design three positional spatial and temporal gating units for video data modeling: \textbf{PoTGU} (temporal), \textbf{PoSGU} (spatial), and \textbf{PoSTGU} (spatio-temporal). To capture various axial relations, they individually take into account the 1D temporal (PoTGU), 2D spatial (PoSGU), and 3D spatio-temporal (PoSTGU) variances. 

\begin{figure*}[t]
\centering
\includegraphics[width=1.0\textwidth]{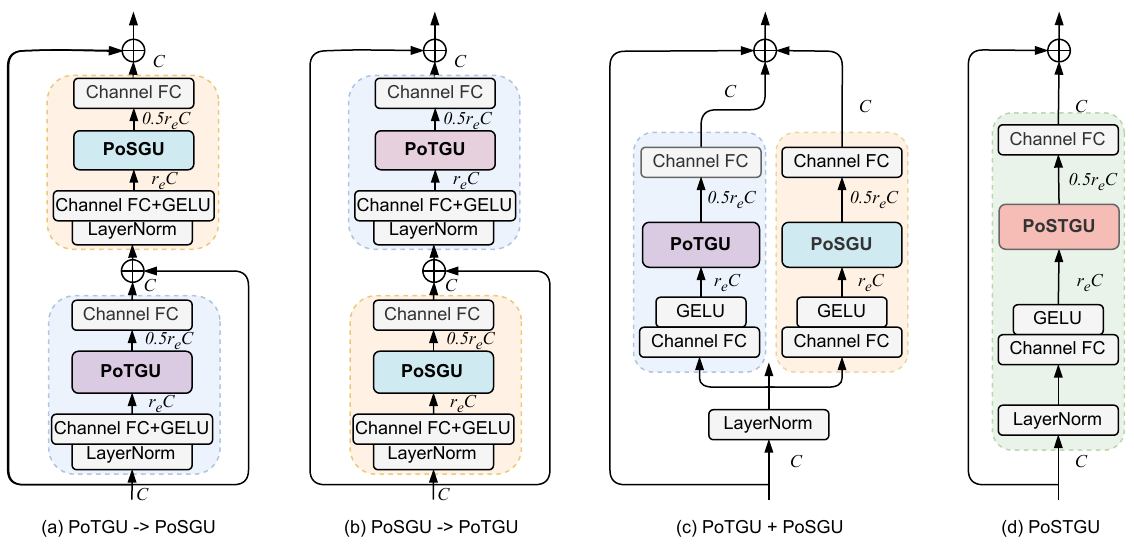}
% \vspace{-0.2cm}
\caption{The schema of the four factorized spatio-temporal PosMLP blocks. The channel expansion ratio $r_e$ is set to 2 and 4 in our implementation.}
\label{model_blocks}
\vspace{-0.3cm}
\end{figure*}

The computation pipeline of the three units follows image PosMLP \cite{wang2022parameterization} and gMLP \cite{liu2021pay}. Formally, we have the general formula as 
\begin{equation}
    \mathbf{Z} = \mathbf{R}\mathbf{X}_{1}\odot \mathbf{X}_2,
\end{equation}
where $\mathbf{X}_{1},\mathbf{X}_2 \in \mathbb{R}^{M^2\times \frac{d}{2}}$ are two independent parts of $\mathbf{X}$ along the channel dimension, $\mathbf{Z} \in \mathbb{R}^{M^2\times \frac{d}{2}}$ is the output with cross-token interactions, and $\odot$ denotes element-wise multiplication. Note that, for the sake of simplicity, we omit the bias term in the equation, as well as in the subsequent ones. Here, $\mathbf{R}$ plays a similar role as the token projection weight matrix of gMLP' spatial gating unit (SGU) that computes token-to-token relations. But, $\mathbf{R}$ further considers the relative position differences between tokens which makes it more suitable for sequential dependency modeling. In the experiment, we verify this by replacing all position units with gMLP's units and observing significant performance drops.

In PosMLP-Video, all position units act on the 3D visual window. Firstly, let's denote the token embeddings of a 3D window as $\mathbf{V}\in \mathbb{R}^{T\times H\times W \times C}$, where $T, H/W, C$ are the window-time, window-height/width and channel dimension, respectively. We first split $\mathbf{V}$ into $\mathbf{V}_{1}\in \mathbb{R}^{T\times H\times W \times \frac{C}{2}}$ and $\mathbf{V}_2\in \mathbb{R}^{T\times H\times W \times \frac{C}{2}}$, and then divide each into $g$ feature groups: $\{\mathbf{V}_{1}^{1},\mathbf{V}_{1}^{2},\cdots,\mathbf{V}_{1}^{g}\} \in \mathbb{R}^{T\times H\times W \times \frac{C}{2g}}$ and $\{\mathbf{V}_{2}^{1},\mathbf{V}_{2}^{2},\cdots,\mathbf{V}_{2}^{g}\} \in \mathbb{R}^{T\times H\times W \times \frac{C}{2g}}$. The channel grouping strategy follows \cite{wang2022parameterization,hao2022group}, which is to increase the multiformity of relative position biases. Below, we present the designs of PoTGU, PoSGU, and PoSTGU in detail.

\textbf{PoTGU} aims to model the 1D temporal relation between tokens, as shown in Figure \ref{model_posv}(a). PoTGU parameterizes $g$ temporal relative position dictionaries $\{\mathbf{P}_{T}^{1},\mathbf{P}_{T}^{2},\cdots,\mathbf{P}_{T}^{g}\}\in \mathbb{R}^{2T-1}$, based on which we build $g$ temporal relative position matrices $\{\mathbf{R}_{T}^{1},\mathbf{R}_{T}^{2},\cdots,\mathbf{R}_{T}^{g}\}\in \mathbb{R}^{T\times T}$ following the LRPE principle. Specifically, for the $i$-th token group (i.e., $i\in [1,2,\cdots,g]$) and $j$-th spatial position (i.e., $j\in [1,2,\cdots,HW]$), the refined output token feature $\mathbf{Z}_T$ is computed as 
\begin{equation}
    \mathbf{Z}_T^{i,j}=\mathbf{R}_{T}^{i}\mathbf{V}_{1}^{i,j}\odot \mathbf{V}_2^{i,j},
\end{equation}
where $\mathbf{V}_{1}^{i,j},\mathbf{V}_2^{i,j},\mathbf{Z}_T^{i,j}$ are with the size of $T\times \frac{C}{2g}$. 
% After computing for each feature group, the final refined 3D token embedding matrix $\mathbf{Z}_T$ is the concatenation of the $g$ refined feature groups along channel dimension, i.e., $\mathbf{Z}_T=\text{Concat}(\mathbf{Z}_T^{1},\mathbf{Z}_T^{2},\cdots,\mathbf{Z}_T^{g})$.

\textbf{PoSGU}, in contrast to PoTGU, pays attention to the 2D spatial relation between tokens, as shown in Figure \ref{model_posv}(b). Similarly, PoSGU learns $g$ spatial relative position dictionaries $\{\mathbf{P}_{S}^{1},\mathbf{P}_{S}^{2},\cdots,\mathbf{P}_{S}^{g}\}\in \mathbb{R}^{(2H-1)\times (2W-1)}$, where there is a total of $(2H-1)\times (2W-1)$ relative position biases in each group. By indexing the position offset, we can construct $g$ spatial relative position matrices $\{\mathbf{R}_{S}^{1},\mathbf{R}_{S}^{2},\cdots,\mathbf{R}_{S}^{g}\}\in \mathbb{R}^{HW\times HW}$. For each frame, the spatial refined token embeddings $\mathbf{Z}_S$ can be calculated as
\begin{equation}
    \mathbf{Z}_S^{i,l}=\mathbf{R}_{S}^{i}\mathbf{V}_{1}^{i,l}\odot \mathbf{V}_2^{i,l},
\end{equation}
where $\mathbf{V}_{1}^{i,l},\mathbf{V}_2^{i,l},\mathbf{Z}_S^{i,l} \in \mathbb{R}^{HW\times \frac{C}{2g}}$, $i\in [1,2,\cdots,g]$ and $l\in [1,2,\cdots,T]$.

\textbf{PoSTGU} treats the 3D feature tensor as $THW$ spatio-temporal tokens and captures their correlations from the spatio-temporal view. Specifically, by employing LRPE, PoSTGU also constructs the $g$ spatio-temporal relative position matrices $\{\mathbf{R}_{ST}^{1},\mathbf{R}_{ST}^{2},\cdots,\mathbf{R}_{ST}^{g}\}\in \mathbb{R}^{THW\times THW}$ based on $g$ learned spatio-temporal relative position dictionaries $\{\mathbf{P}_{ST}^{1},\mathbf{P}_{ST}^{2},\cdots,\mathbf{P}_{ST}^{g}\}\in \mathbb{R}^{(2T-1)\times(2H-1)\times (2W-1)}$. The final refined token embedding matrix $\mathbf{Z}_{ST}$ is hereby computed as
\begin{equation}
    \mathbf{Z}_{ST}^{i}=\mathbf{R}_{ST}^{i}\mathbf{V}_{1}^{i}\odot \mathbf{V}_2^{i},
\end{equation}
where $\mathbf{V}_{1}^{i},\mathbf{V}_2^{i},\mathbf{Z}_{ST}^{i} \in \mathbb{R}^{THW\times \frac{C}{2g}}$, and $i\in [1,2,\cdots,g]$. Figure \ref{model_posv}(c) shows the pipeline. 

\begin{figure*}[t]
\centering
\includegraphics[width=1.0\textwidth]{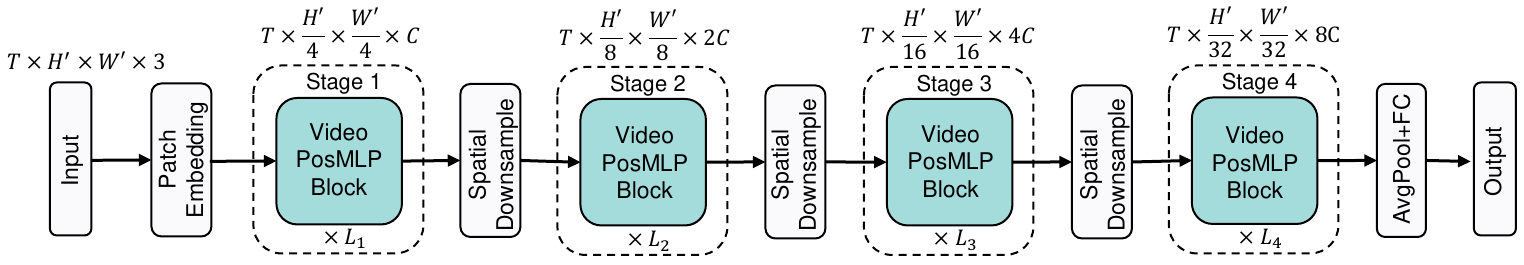}
% \vspace{-0.2cm}
\caption{Overall architecture of PosMLP-Video. }
\label{model_arch}
% \vspace{-0.3cm}
\end{figure*}

\textbf{Complexity Comparison}. We list the parameters counting of gMLP's SGU and our positional units in Table \ref{tab:compcost}. It can be found that the proposed PoTGU/PoSGU/PoSTGU have much fewer parameters than SGU. For example, by setting the 3D window size as $\{T=16, H=7, W=7\}$ and $g=8$, PoTGU, PoSGU and PoSTGU have 248, 1,352 and 41,912 parameters compared to 615,440 of SGU. In the comparison with self-attention, since there is no pairwise computation in position units, our PosMLP-Video also shows much lower FLOPs than video Transformers as demonstrated by the experiment. 
% Also, since the construction of relative position matrix $\mathbf{R}$ is only an indexing operation, it requires extremely low FLOPs as demonstrated by the experiment. 
\begin{table}[h]
            \caption{Comparison of parameters between gMLP's SGU and our position variants. Note that we omit the bias term.}
		\label{tab:compcost}
		\centering
        % \footnotesize
		\scriptsize
		% \small
		\begin{tabular}{l|c}
		% \hline
        \toprule[1pt]
	     \textbf{Module} &\textbf{Parameters} \\
            \midrule[1pt]
            % \hline
            SGU (gMLP)   &$THW\times THW$  \\
            \hline
            PosTGU       &$g\times(2T-1)$  \\
            % \hline
            PosSGU       &$g\times(2H-1)\times(2W-1)$   \\
            % \hline
            PosSTGU      &$g\times(2T-1)\times(2H-1)\times(2W-1)$   \\
            \bottomrule[1pt]
		\end{tabular}
		% \vspace{-0.2cm}
		
	% \vspace{-0.2cm}
\end{table}

\subsection{Video PosMLP Blocks}
PoTGU/PoSGU/PoSTGU model the cross-token interaction. To further enhance the cross-channel interaction, we follow gMLP \cite{liu2021pay} that adds channel FC layer before and after the pos gating unit, resulting in three corresponding PosMLP modules. In practice, they have a similar network structure, as shown in Figure \ref{model_blocks}. Specifically, we first add a LayerNorm and a channel FC layer followed by GELU activation before the pos unit. The channel is expanded with a ratio $r_e$ in this FC layer. Then another channel FC layer is put after the pos unit to reshape the channel size to $C$. 

To achieve the modeling of spatial-temporal interactions among the tokens in a 3D view, we introduce three factorized spatio-temporal PosMLP blocks by combining PoTGU and PoSGU, including two cascaded blocks: \textbf{PoTGU$\rightarrow$PoSGU} and \textbf{PoSGU$\rightarrow$PoTGU}, and one paralleled block: \textbf{PoTGU+PoSGU}. As a comparison, we also present the joint spatio-temporal PosMLP block by using \textbf{PoSTGU}. Their architectural designs are shown in Figure \ref{model_blocks}. Particularly, the factorized spatio-temporal design shares a similar spirit with the existing works in video CNNs \cite{qiu2017learning,hao2022group,xie2018rethinking}, Transformers \cite{arnab2021vivit,bertasius2021space} and MLPs \cite{zhang2022morphmlp,qiu2022mlp}. As demonstrated in their experiments as well as ours, the factorized architectures can achieve a preferable balance between accuracy and efficiency.

\subsection{PosMLP-Video Architecture}\label{archi}

Figure \ref{model_arch} illustrates the overall architecture. Particularly, the patch embedding block at the top of the model receives the input of a raw video and outputs token embeddings. The spatial downsampling layers between the four stages are to reduce the spatial resolution with a ratio of 2. Following \cite{wang2022parameterization}, we utilize the spatial window partitioning strategy on stages 1-4 to produce multiple non-overlapping token embedding windows, on which the video PosMLP blocks act. The used window sizes are $14\times 14$ for stages 1-3 and $7\times 7$ for the last stage 4. Two obvious advantages of using window partitioning are: (1) small windows enhance the locality of modeling \cite{liu2021swin,liu2022video}; and (2) parameters of a video PosMLP block are shared among windows, significantly reducing the model complexity.

We provide three PosMLP-Video variants, including PosMLP-Video-S (small), PosMLP-Video-B (base) and PosMLP-Video-L (large). Their architecture settings are listed in Table \ref{tab:sbl}. Their differences lie in $r_e$ and layer numbers of stages 1-4. It is worth noting that the temporal length $T$ remains unchanged throughout the network, which is different from MorphMLP and MLP-3D that reduce $T$ to $\frac{T}{2}$ after patch embedding. We find this setting can result in much better performance.

The patch embedding module consists of two standard convolution layers with the kernel (1,3,3) and stride (1,2,2). A batchnorm layer (BatchNorm3D) is added after each convolution. The activation GELU is inserted between the two convolution layers. The spatial downsampling module is implemented as a standard convolution layer with the kernel (1,3,3) and stride (1,2,2) followed by a LayerNorm. The two modules are the same as MorphMLP. To further evaluate the impact of patch embedding, we conduct experiments using three different designs that explore the interactions between patch neighbors.

\begin{table*}[htbp]
\caption{Architecture settings of MorphMLP, MLP-3D and our PosMLP-Video models.}\label{tab:sbl}
		\centering
		% \scriptsize
		% \small
        \footnotesize
        % \resizebox{1.0\textwidth}{!}{
		\begin{tabular}{l|c|c|c}
		% \
        \toprule[1pt]
%         \cline{7-9}\cline{10-11}
	     \multirow{2}*{\textbf{Models}} &\multirow{2}*{$T\text{(input)}\rightarrow T'$(after patch emb.)}  &\textbf{Layer numbers} &\textbf{Layer channels} \\
           &&$\{L_1,L_2,L_3,L_4\}$   &$\{C_1,C_2,C_3,C_4\}$ \\
            % \midrule[1pt]
            \hline
            PosMLP-Video-S ($r_e$=2) &\multirow{3}*{16$\rightarrow16$}  &$\{3,4,9,3\}$ &\multirow{3}*{$\{72, 144, 288, 576\}$} \\
            % \cline{1-2}
            PosMLP-Video-B ($r_e$=2) & &$\{4,6,15,4\}$ & \\
            % \cline{1-2}
            PosMLP-Video-L ($r_e$=4) & &$\{4,6,15,4\}$ & \\
            \midrule[1pt]
            MorphMLP-S  &\multirow{2}*{16$\rightarrow8$} &$\{3,4,9,3\}$  &\multirow{2}*{$\{112, 224, 392, 784\}$} \\
            % \cline{1-2}
            MorphMLP-B & &$\{4,6,15,4\}$ & \\
            \midrule[1pt]
            MLP-3D-S &\multirow{3}*{16$\rightarrow8$} &$\{2,3,10,3\}$ &\multirow{2}*{$\{64, 128, 320, 512\}$} \\
            % \cline{1-2}
            MLP-3D-M & &$\{3,4,18,3\}$ & \\
            % \hline
            % \cline{1-2}
            MLP-3D-L & &$\{3,2,24,3\}$ & $\{96, 192, 384, 768\}$ \\
            % \hline
            \bottomrule[1pt]
    		\end{tabular}
      % }
\end{table*}

Our PosMLP-Video can also benefit from the pretraining on large-scale image datasets like ImageNet1K. Since only the PoTGU and PoSTGU need to perform on the time axis, we can simply replace them with a residual connection to facilitate image modeling. When dealing with videos, the only parameters, i.e., the relative position biases, of PoTGU and PoSTGU will be randomly initialized.

\subsection{Discussion} 
Here, we present a comprehensive discussion between our PosMLP-Video and two other MLP-based video neural networks, MorphMLP~\cite{zhang2022morphmlp} and MLP-3D~\cite{qiu2022mlp}. All three models share a similar hierarchical framework with four stages but differ in their MLP block designs within each stage.

Specifically, the MorphMLP block comprises two sequential layers: the spatial $\text{MorphFC}_s$ and the temporal $\text{MorphFC}_t$. $\text{MorphFC}_s$ splits the spatial feature into chunks along the horizontal or vertical direction, flattens each chunk into a 1D vector, and applies an FC weight matrix for transformation on each chunk. $\text{MorphFC}_t$ concatenates features across all frames, creating a chunk, and uses an FC matrix to model temporal relationships. In summary,  both $\text{MorphFC}_s$ and $\text{MorphFC}_t$ utilize an FC layer on flattened chunk vectors to model spatial and temporal token-wise relations. On the other hand, the MLP-3D block consists of a decomposed token-mixing MLP and a channel MLP. The decomposed token-mixing MLP divides the original token-mixing MLP in \cite{tolstikhin2021mlp} into height, width, and grouped time mixing variants, with its output representing the weighted summation of these three sub-operations. In essence, MLP-3D introduces a space-time decomposed token-mixing MLP for modeling spatio-temporal token relations.

In contrast, our PosMLP block in PosMLP-Video utilizes learnable relative position encoding to model token relations in both spatial and temporal axes, which sets our approach apart from MorphMLP and MLP-3D. Additionally, both MorphMLP and MLP-3D employ channel grouping operations, although their motivations differ from ours. MorphMLP and MLP-3D use channel grouping to facilitate weight-sharing across different feature groups, while our PosMLP aims to enhance token relationships by utilizing multiple groups with each one learning a specific relative position dictionary.

Regarding model parameters and computations, both the channel FC layers and spatio-temporal operations contribute to the overall count. Since the three methods have similar channel FC layers, the differences in model size primarily arise from the spatio-temporal operations.  In comparison to MorphMLP and MLP-3D, which employ FC projections with $O(N^2)$ complexity, our PosMLP-Video utilizes LRPE to learn relative position dictionaries with only  $O(N)$ complexity. The architectural hyperparameters of the three models are listed in Table \ref{tab:sbl}. PosMLP-Video adopts the same stage settings as MorphMLP, but the three models differ in their feature channel settings across the four stages. In our experiments, we also conduct tests using the same channel settings as the other two models to facilitate a more comprehensive comparison.

\section{Experiment on Video Recognition}
We examine our PosMLP-Video models mainly on video classification tasks, for both coarse-grained and fine-grained video actions. Following prior art, top-1 and top-5 accuracies (\%) are adopted to evaluate the performance. Parameters and FLOPs are also reported to show the model complexity.

\subsection{Datasets}
We use 5 standard video benchmark datasets, including Kinetics-400 (K400) \cite{kay2017kinetics}, Something-Something V1 (SSV1) and V2 (SSV2) \cite{goyal2017something}, Diving48 \cite{li2018resound} and  EGTEA Gaze+ \cite{li2018eye}. \textbf{Kinetics-400} is a large-scale video dataset, containing $\sim$246k/20k training/validation videos for 400 human action classes. The actions in Kinetics-400 are relatively coarse and prefer spatial context to temporal context. \textbf{Something-Something V1} and \textbf{V2} cover 174 fine-grained human performing activities and require temporal modeling more. Particularly, V1 is the smaller version and has $\sim$86K/12K training/validation videos, while the larger V2 contains $\sim$169k/25k videos. \textbf{Diving48} is another fine-grained video dataset, consisting of $\sim$18k trimmed video clips of 48 unambiguous dive sequences. Here, the newly released dataset version (V2) is used. \textbf{EGTEA Gaze+} is a first-person video dataset covering 106 fine-grained daily action categories, where the split-1 that contains $\sim$8.3k/3.8k training/validation clips is selected for use.

\subsection{Implementation Details}
The implementation of PosMLP-Video variants is built upon the PySlowFast \cite{fan2020pyslowfast} repository and mostly follows MViT \cite{fan2021multiscale} and MorphMLP \cite{zhang2022morphmlp} for training and validation protocols. All experiments are run on servers with 4$\times$3090 or 4$\times$A100 GPUs.

\textbf{Pretraining.} As explained in Section \ref{archi}, our PosMLP-Video variants can be pre-trained on the image dataset. The pertaining settings follow \cite{wang2022parameterization}. After obtaining the pertained weights, we can easily use them to initialize most of the video model layers and blocks except for the relative position biases of PoTGU and PoSTGU. In the experiment, we adopt ImageNet1K for pertaining and observe comparable or even better performance than those pre-trained on the larger-scale ImageNet21K.

\textbf{Training.} For all the five datasets, each frame of a video is firstly resized to $256\times320$ and then cropped to $224\times224$ as model input. The temporal length $T$ is set to 16/24. Particularly, for Kinetics-400, the dense sampling strategy is adopted to select $T$ video frames, and the training configurations are set as follows: warm-up epoch 10, total epoch 60, batch size 8 per GPU, and base learning rate 2e-4 and weight decay 0.05 for AdamW optimizer. The random horizontal flip is also adopted. Stochastic depth rates are set to 0.05/0.1/0.2 for S/B/L. For other datasets, the sparse sampling strategy is used. Most of the training settings are the same with Kinetics-400, except warm-up epoch 5 and base learning rate 4e-4. Here, stochastic depth rates are set to 0.1/0.3/0.5 for S/B/L. 

\textbf{Inference.} For Kinetics-400, we uniformly sample four clips from each test video with three crops \cite{feichtenhofer2019slowfast}. While, for other datasets, we only extract one clip with each having one or three crops. The form of ``$A\times B\times C$'' denotes $A$ frames ($T$), $B$ crops and $C$ clips in the tables.

\begin{table}[htbp]
\caption{Performance comparison with different positional gating units on SSV1 dataset. The results are obtained without pertaining.}
\label{tab:posvariant}
		\centering
		% \scriptsize
        \footnotesize
% 		\small          
        % \resizebox{1.02\columnwidth}{!}{
		\begin{tabular}{l|cc|cc}
		\toprule
		\textbf{Positional Gating Unit} &\textbf{Params} &\textbf{GFLOPs} &\textbf{Top-1} &\textbf{Top-5} \\
            \midrule[1pt]
            % \hline
            PoSGU  &7.95M &25.08 &6.25 &20.64 \\
            PoTGU  &7.65M  &20.32 &25.44 &51.36 \\
            PoSTGU  &17.19M  &103.09 &40.89 &70.55 \\
            \bottomrule
		\end{tabular}
  % }
\end{table}

\begin{table}[htbp]
\caption{Performance comparison with different video PosMLP blocks on SSV1 dataset. The results are obtained without pertaining except the last one. ``IN-1K'' means that the model is pretrained on ImageNet1K.}
\label{tab:posblock}
		\centering
		% \scriptsize
        \footnotesize
% 		\small          
        % \resizebox{1.02\columnwidth}{!}{
		\begin{tabular}{l|cc|cc}
		\toprule
		\textbf{Video PosMLP Block} &\textbf{Params} &\textbf{GFLOPs} &\textbf{Top-1} &\textbf{Top-5} \\
            \midrule[1pt]
            % \hline
            PoTGU$\rightarrow$PoSGU   &13.51M  &40.49 &44.11 & 74.70 \\
            PoSGU$\rightarrow$PoTGU   &13.51M  &40.49 &42.40 & 72.36 \\
            PoTGU+PoSGU   &13.51M  &40.49 &\textbf{46.31} &\textbf{75.34} \\
            % SGU+TGU (gMLP)  &13.83M  &40.76 &42.94 &72.17 \\
            \hline
            PoTGU+PoSGU (IN-1K)   &13.51M  &40.49 &\textbf{52.24} &\textbf{78.96} \\
            \bottomrule
		\end{tabular}
  % }
\end{table}

\begin{table}[htbp]
\caption{Performance comparison with different video blocks on SSV1 dataset. The results are obtained without pertaining. ``TConv'' denotes the temporal convolution using a 3 kernel, while ``SConv'' denotes the spatial convolution using $3\times 3$ kernel. ``TShift'' is the temporal shift operation based on TSM~\cite{lin2019tsm}. ``SAttention'' refers to the spatial self-attention. TGU is a temporal version of SGU in gMLP.}
\label{tab:otherstoperation}
		\centering
		% \scriptsize
        \footnotesize
% 		\small          
        % \resizebox{1.02\columnwidth}{!}{
		\begin{tabular}{l|cc|cc}
		\toprule
		\textbf{Video Block} &\textbf{Params} &\textbf{GFLOPs} &\textbf{Top-1} &\textbf{Top-5} \\
            \midrule[1pt]
            % \hline
            \textbf{PoTGU+PoSGU (our)}   &13.51M  &40.49 &\textbf{46.31} &\textbf{75.34} \\
            \hline
            PoTGU+SConv        &24.23M  &64.94 &37.96 &66.53 \\
            PoTGU+SAttention   &15.25M  &42.93 &30.42 &58.80 \\
            TConv+PoSGU        &13.48M  &40.04 &42.25 &72.17 \\
            TShift+PoSGU       &7.95M  &25.08 &41.84 &71.94 \\
            SGU+TGU (gMLP)     &13.83M  &40.76 &42.94 &72.17 \\
            \bottomrule
		\end{tabular}
  % }
\end{table}

\begin{table*}[t]
\caption{Performance comparison with different window sizes (time length ($T$)$\times$space ($H\times W$)) on SSV1 dataset. The used $g$ is $(8,16,32,64)$ for Stage 1-4. The results are obtained without pretraining.}
		\label{tab:diffwindow}
		\centering
		% \scriptsize
        % \footnotesize
		\small
        % \resizebox{1.0\columnwidth}{!}{
		\begin{tabular}{c|cc|c}
			\toprule
			\textbf{Window size Stage (S) 1-4} &\textbf{Params} &\textbf{GFLOPs} &\textbf{Top-1}  \\ %&\textbf{Top-5}
            \midrule[1pt]
            % \hline
            $16\times7\times7$ (S1-4)  &13.30M  &36.64 &44.21 \\ % &73.22
            $16\times14\times14$ (S1-3),$16\times7\times7$ (S4)   &13.51M  &40.49 &\textbf{46.31}  \\ %&75.34
            $8\times14\times14$ (S1-3),$8\times7\times7$ (S4)   &13.50M  &40.27 &40.73 \\ %&70.11
            $16\times28\times28$ (S1), $16\times14\times14$ (S2-3),$16\times7\times7$ (S4)   &13.57M  &46.86 &46.29  \\ %&75.58
            \bottomrule
		\end{tabular}
  % }
		% \vspace{-0.1cm}
\end{table*}

\begin{table}[h]
\caption{Performance comparison with different group numbers $g$ on SSV1 dataset. The used window size is the default setting. The results are obtained without pretraining.}
		\label{tab:diffgroup}
		\centering
		% \scriptsize
        % \footnotesize
		% \small
        % \resizebox{0.95\columnwidth}{!}{
		\begin{tabular}{c|c|cc|c}
			\toprule
			\textbf{Model} &\textbf{$g$ Stage 1-4} &\textbf{Params} &\textbf{GFLOPs} &\textbf{Top-1} \\
            \midrule[1pt]
            % \hline
            \multirow{5}*{Small} &$(1,1,1,1)$  &13.20M  &40.49 &43.90  \\
            &$(8,8,8,8)$  &13.24M  &40.49 &45.62  \\
            &$(4,8,16,32)$  &13.35M  &40.49 &44.87  \\
            &$(8,16,32,64)$  &13.51M  &40.49 &\textbf{46.31}  \\
            &$(12,24,48,96)$  &13.67M  &40.49 &45.96  \\
            \hline
            \multirow{2}*{Base} &$(8,16,32,64)$  &18.98M  &58.93 &\textbf{47.33}  \\
             &$(12,24,48,96)$  &19.24M  &58.93 &47.21  \\
            \bottomrule
		\end{tabular}
  % }
\end{table}

\begin{table*}[h]
\caption{Performance comparison with different channel numbers on SSV1 dataset. The used window size is the default setting. All the models pretrained on IN-1K. The inference is based on a single clip.}
		\label{tab:diffchannel}
		\centering
		% \scriptsize
        \footnotesize
		% \small
        % \resizebox{0.99\columnwidth}{!}{
		\begin{tabular}{c|l|c|cc|c}
			\toprule
			\textbf{Model} &$(C_1, C_2, C_3, C_4))$  &$T$(input)$\rightarrow T'$(after patch emb.) &\textbf{Params} &\textbf{GFLOPs} &\textbf{Top-1}  \\
            \midrule[1pt]
            % \hline
            \multirow{5}*{\textbf{PosMLP-Video-S}} &$(64,128,320,512)$ & \multirow{4}*{$16\rightarrow16$}  &13.12M  &39.70 &50.02  \\
            &\textbf{$(72,144,288,576)$ (used)}  &&13.51M  &40.49 &52.24  \\
            &$(96,192,384,768)$  &&23.69M  &69.24 &52.34  \\
            &$(112,224,392,784)$  &&25.31M  &80.64 &52.83 \\
            \Xcline{2-6}{0.4pt}
            &$(112,224,392,784)$ &$16\rightarrow8$  &25.30M  &40.46 &51.40  \\
            \hline
            MorphMLP-S~\cite{zhang2022morphmlp} &$(112,224,392,784)$ &$16\rightarrow8$ &46.9M   &67.00 &50.6 \\
            \bottomrule
		\end{tabular}
  % }
\end{table*}

\begin{table*}[h]
		\caption{Performance comparison of different patch embedding modules using PosMLP-Video-S on SSV1 dataset. The results are obtained without pertaining.}
		\label{tab:patch}
		\centering
		% \scriptsize
        % \footnotesize
		\small
        % \resizebox{1.02\columnwidth}{!}{
		\begin{tabular}{l|cccc|cc}
			\toprule
		\textbf{PE Version} &\textbf{Structure}  &\textbf{Patch Overlap} &\textbf{Params} &\textbf{GFLOPs} &\textbf{Top-1} &\textbf{Top-5} \\
            \midrule[1pt]
            % \hline
           PE-V1 &Conv(1,4,4) & \usym{2718} &13.49M  &39.28 &45.41 &75.00 \\
           PE-V2 &Conv(1,2,2)$\rightarrow$Conv(1,2,2)  &\usym{2718} &13.50M   &39.73 &45.17 &75.07 \\
           PE-V3 (used) &Conv(1,3,3)$\rightarrow$Conv(1,3,3)  &\usym{2714} &13.51M  &40.49  &\textbf{46.31} &\textbf{75.34} \\
            \bottomrule
		\end{tabular}
  % }
\end{table*}

\begin{table}[h]
		\caption{Performance comparison of GQPE and LRPE using PosMLP-Video-S on SSV1 dataset. The results are obtained without pertaining.}
		\label{tab:gqpe}
		\centering
		% \scriptsize
        % \footnotesize
% 		\small
        % \resizebox{1.02\columnwidth}{!}{
		\begin{tabular}{l|cc|cc}
			\toprule
			\textbf{RPE method} &\textbf{Params} &\textbf{GFLOPs} &\textbf{Top-1} &\textbf{Top-5} \\
            \midrule[1pt]
            % \hline
            GQPE  &13.19M &40.56 &38.06 &67.45 \\
            LRPE (used)   &13.51M  &40.49 &\textbf{46.31} &\textbf{75.34} \\
            % PoTGU+PoSGU (Pretain on IN-1K)   &13.51M  &40.49 &\textbf{52.24} &\textbf{78.96} \\
            \bottomrule
		\end{tabular}
  % }
\end{table}

\subsection{Ablation Study}
In the ablation study, we thoroughly investigate various hyperparameters, settings, and model design variations to comprehensively address the design questions raised in the introduction. All the results are obtained on the Something-Something V1 dataset and using 16 frames as input.

{\bf Positional spatial and temporal gating units.} We examine the performance of the three proposed gating units PoSGU, PoTGU and PoSTGU. The results are shown in Table \ref{tab:posvariant} and the spatial window size is set as $14\times 14$ for stages 1-3 and $7\times 7$ for stage 4 by following the work~\cite{wang2022parameterization}. It can be found that the single temporal PoTGU can obtain higher top-1 accuracy than the single spatial PoSGU, while the spatio-temporal PoSTGU surpasses PoTGU and PoSGU. Since categorizing videos in SSV1 requires specific spatio-temporal relation modeling, this result trend is expected. This observation verified that the proposed temporal expansion for the positional spatial gating unit of image PosMLP can successfully capture time information in a video.

{\bf Video PosMLP block variants.} Although the proposed positional saptio-temporal gating unit PoSTGU can achieve much better top-1 accuracy than PoSGU and PoTGU. Its model parameters and FLOPs are quite larger than the others as shown in Table \ref{tab:posvariant}. By considering this, we propose three combination versions of PoSGU and PoTGU using space-time factorization method, resulting in two cascaded blocks: PoTGU$\rightarrow$PoSGU and PoSGU$\rightarrow$PoTGU, and one paralleled block: PoTGU+PoSGU. Table \ref{tab:posblock} shows the results of the three model variants. Among them, the paralleled PoTGU+PoSGU achieves the best performance. In the last line, we also provide the results of PoTGU+PoSGU with pre-training on ImageNet1K, improving the top-1 accuracy of w/o pre-training from 46.31\% to 52.24\%.

In addition, we also conduct experiments by replacing the positional units with other operators such as convolution, temporal shift, and token-mixing-based gating units. The results are shown in Table \ref{tab:otherstoperation}. We can observe that our PoSGU and PoTGU significantly outperform all the other operations with the used structural framework.

{\bf Different window size.} Window partitioning is used to split the whole spatio-temporal video clip into several non-overlapped parts, whose effect has been demonstrated by Swin Transformers \cite{liu2021swin,liu2022video}. For the spatial window size, the image PosMLP provides a good choice for the setting, i.e., $14\times 14$ for stages 1-3 and $7\times 7$ for stage 4. In the video processing, we further test different window sizes for performance change observation. we find that (1) temporal partitioning, e.g., $T=8$ compared with $T=16$,  leads to noticeable performance degradation (40.73\% vs. 46.31\%), and (2) larger spatial window size results in more parameters and FLOPs while performance does not show monotonic increasing. In other words, the large temporal receptive field is more important for action recognition on SSV1. By considering the complexity-accuracy trade-off, we select $16\times14\times14$ for Stage 1-3 and $16\times7\times7$ for Stage 4 as the ultimate window size settings.

{\bf Different group number $g$ for multiple relative position bias dictionaries.} We examine the impact of group number $g$. We can find in Table \ref{tab:diffgroup} that increasing $g$ will monotonically raise the model size (parameters) while boosting performance significantly. This is mainly because that larger $g$ leads to more relative position bias dictionaries and also enlarges the model capacity for diverse spatio-temporal relations. However, the performance does not increase without bound. When increasing $g$ to $(12,24,48,96)$, the performance is relatively lower than that with $(8,16,32,64)$.  Finally, considering the trade-off between model size and accuracy, we set $g$ to $(8,16,32,64)$ for stages 1-4. Note that it differs from convolution in that group operation will reduce the model size. This is because the proposed positional gating unit does not involve channel interactions.

{\bf Different channel numbers.} As shown in Table \ref{tab:sbl} in Section \ref{archi}, our PosMLP-Video adopts distinct channel configurations compared to MorphMLP and MLP-3D. Additionally, both MorphMLP and MLP-3D reduce the time length from $T$ to $\frac{T}{2}$ after patch embedding. In our experiments, we not only examine channel settings identical to MorphMLP and MLP-3D but also implement a $T\rightarrow\frac{T}{2}$ (i.e., $16\rightarrow8$) transformation in our PosMLP-Video-S variant for a more precise comparison with MorphMLP-S, as detailed in Table \ref{tab:diffchannel}. Firstly, it's observed that higher channel numbers generally yield improved top-1 accuracies, but they come at the cost of significantly increased model size and computational complexity. Secondly, highlighted in the last two rows of the table, the reduction of temporal length $T\rightarrow\frac{T}{2}$ substantially lowers the computational cost (e.g., from 80.64G to 40.46G FLOPs). However, this reduction is accompanied by a ~1.4\% decrease in top-1 accuracy. Based on these findings,  we set the ultimate channel setting as $(72,144,288,576)$ to strike a favorable balance between model complexity and accuracy.

\begin{figure}[]
\centering
\includegraphics[width=0.49\textwidth]{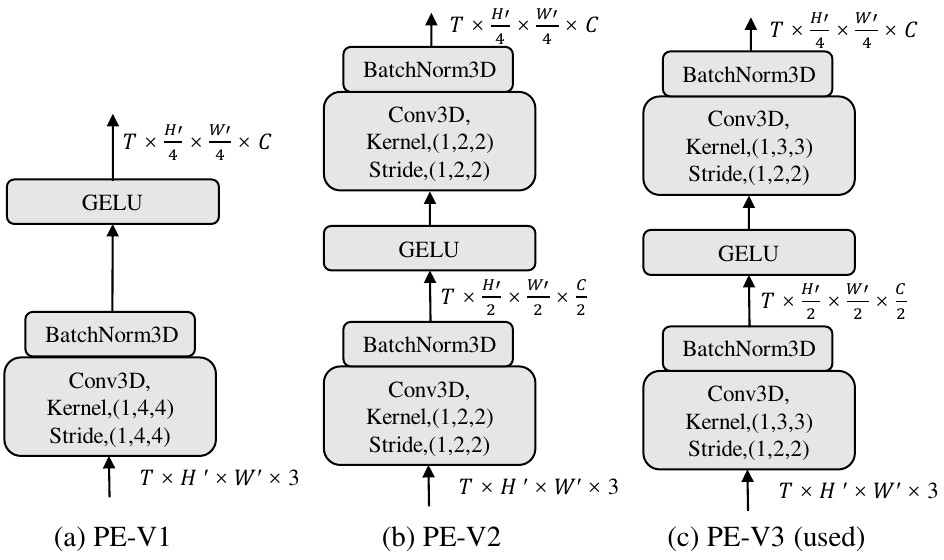}
% \vspace{-0.2cm}
\caption{Patch embedding modules.}
\label{patchemb}
% \vspace{-0.3cm}
\end{figure}

{\bf Patch embedding designs.} Patch embedding (PE) is a crucial step in visual Transformers. The conventional PE structure, seen in Transformers like ViT and Swin-Transformer, is depicted in Figure \ref{patchemb}(a), referred to as PE-V1. It consists of a single convolution layer with $(1,4,4)$ kernel and $(1,4,4)$ stride, dividing the image into patches. Consequently, there is no direct interaction between adjacent patches. We explore two additional PE designs by employing two cascaded convolution layers: (1) each layer with $(1,2,2)$ kernel and $(1,2,2)$ stride, referred to as PE-V2 (Figure \ref{patchemb}(b)) and (2) each layer with $(1,3,3)$ kernel and $(1,2,2)$ stride, referred to as PE-V3 (Figure \ref{patchemb}(c)). PE-V3 achieves patch overlap by using a larger convolution kernel and a smaller stride compared to PE-V1 and PE-V2. Table \ref{tab:patch} compares their results. Notably, patch overlap during patch embedding learning yields better results than the non-overlapping approach.

{\bf  LRPE vs. GQPE.} In addition to LRPE, we also test the performance of another RPE method called GQPE. Table \ref{tab:gqpe} compares the performance of both methods, and it was observed that LRPE significantly outperforms GQPE. Despite having fewer parameters, with the model complexity of GQPE being $O(1)$ compared to $O(N)$ of LRPE, the model capacity of GQPE is probably constrained. This is due to the fact that the RPE-based relation score in GQPE is determined solely by the relative position and is not learnable.

\begin{figure*}[]
\centering
\includegraphics[width=1.0\textwidth]{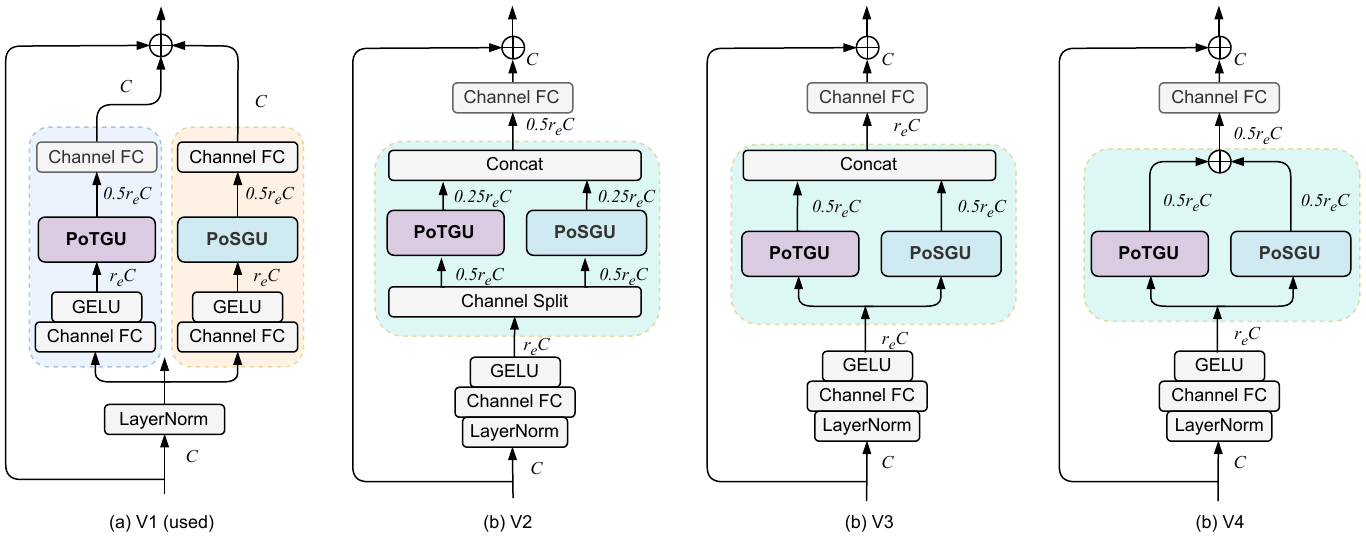}
% \vspace{-0.2cm}
\caption{PoTGU+PoSGU versions. ``V1'' is the used version of PosMLP-Video. ``V2'' adopts the channel splitting before inputting into the pos units and then concatenates the outputs of PoTGU and PoSGU along the channel dimension. ``V3''  separately inputs the feature into PoTGU and PoSGU and then concatenates their outputs along the channel dimension. In contrast to V3, ``V4''  elementwisely adds the outputs of PoTGU and PoSGU.}
\label{model_posblockversion}
% \vspace{-0.3cm}
\end{figure*}

\begin{table}[ht]
        \caption{Performance comparison of different PoTGU+PoSGU versions using PosMLP-Video-S on SSV1 dataset. The results are obtained without pertaining.}
		\label{tab:posblockversion}
		\centering
		% \scriptsize
        \footnotesize
		% \small
        % \resizebox{1.02\columnwidth}{!}{
		\begin{tabular}{c|l|c|cc|c}
			\toprule
		\textbf{Model} &\textbf{Variant} &\textbf{$r_e$} &\textbf{Params} &\textbf{GFLOPs} &\textbf{Top-1} \\
            \midrule[1pt]
            % \hline
           \multirow{4}*{Small} &V1 (used)   &2 &13.51M  &40.49 &\textbf{46.31} \\
            &V2          &4 &13.49M  &40.35 &45.92  \\
            &V3          &2 &9.80M  &30.46 &43.98  \\
            &V4          &2 &7.96M  &25.52 &42.52  \\
            \midrule[1pt]
            \multirow{2}*{Base} &V1 (used)   &2 &18.98M  &58.93 &\textbf{47.33}  \\
            &V2          &4 &18.96M  &58.73 &47.28  \\
            \bottomrule
		\end{tabular}
  % }
\end{table}

\begin{table*}[]
\caption{Comparison of performance on Something-Something V1 dataset.}
		\label{tab:res_somev1}
		\centering
		% \scriptsize
        % \footnotesize
		% \small
        \resizebox{1.9\columnwidth}{!}{
		\begin{tabular}{l|c|c|c|c|cc}
			% \hline
   \toprule
% 			& & & & && \multicolumn{2}{c}{V1} && \multicolumn{2}{c}{V2}\\
%         \cline{7-9}\cline{10-11}
			\textbf{Method} &\textbf{Pretrain} &\textbf{Params} &\textbf{Fr.$\times$Cr.$\times$Cl.} &\textbf{GFLOPs} &\textbf{Top-1} &\textbf{Top-5} \\
            \Xcline{1-7}{0.6pt}
            \multicolumn{7}{c}{\textbf{\textit{CNNs}}} \\
            \hline
            TANet-R50 \cite{liu2021tam} &\multirow{13}*{IN-1K} &25.6M &16$\times$1$\times$1 &66 &47.6 &77.7 \\
            TSM \cite{lin2019tsm} & &23.9M  &16$\times$3$\times$2 &197.4  &48.4 &78.1  \\
            SmallBig \cite{li2020smallbignet} & &---  &16$\times$3$\times$2 &--- &50.0 &79.8  \\
            STM \cite{jiang2019stm} &  &24.0M  &16$\times$3$\times$10 &999  &50.7 &80.4  \\
            TEINet \cite{liu2020teinet} & &30.4M  &16$\times$3$\times$10 &1980 &51.0 &---  \\
            AIA(TSM)~\cite{hao2022attention} & &23.9M  &16$\times$3$\times$2 &397.8 &51.6 &79.9  \\
            MSNet \cite{kwon2020motionsqueeze} & &24.6M &16$\times$1$\times$1 &67  &52.1 &82.3 \\
            TEA \cite{li2020tea}   & &--- &16$\times$3$\times$10 &2100 &52.3 &81.9  \\
            SDA-TSM~\cite{tan2021selective} & &25.8M  &16$\times$3$\times$2 &406.8 &52.8 &81.3  \\
            CT-NET~\cite{li2021ct}    & &---  &16$\times$3$\times$2 &447 &53.4 &81.7  \\
            TDN \cite{wang2021tdn} & &26.1M  &16$\times$1$\times$1 &72.0 &53.9 &82.1  \\
            GC-TDN \cite{hao2022group} & &27.4M  &16$\times$1$\times$1 &73.4 &\textbf{55.0} &82.3  \\
            H$^{2}$CN~\cite{tan2022hierarchical} & &24.1M  &16$\times$1$\times$1 &67.6 &\textbf{55.0} &\textbf{82.4}  \\
            \Xcline{1-7}{0.6pt}
            \multicolumn{7}{c}{\textbf{\textit{MLPs}}} \\
            \hline
            MLP-3D-S \cite{qiu2022mlp} &\multirow{6}*{IN-1K} &74.1M &64$\times$3$\times$1  &324 &55.2 &83.2 \\
            MLP-3D-B \cite{qiu2022mlp} & &88.3M &64$\times$3$\times$1  &549 &56.2 &83.5 \\
            MLP-3D-L \cite{qiu2022mlp} & &149.4M &64$\times$3$\times$1  &1008 &56.5 &83.5 \\
            % MorphMLP-S \cite{zhang2022morphmlp} & & &16$\times$3$\times$1 &201  &53.9  &81.3 \\
            MorphMLP-S \cite{zhang2022morphmlp} & &46.9M &16$\times$3$\times$1 &201  &53.9  &81.3 \\
            MorphMLP-B \cite{zhang2022morphmlp} & &67.6M &16$\times$3$\times$1 &294  &55.5  &82.4 \\
            MorphMLP-B \cite{zhang2022morphmlp} & &68.5M &32$\times$3$\times$1 &591  &57.4  &84.5 \\
            \Xcline{1-7}{0.6pt}
            \textbf{PosMLP-Video-S} &\multirow{3}*{IN-1K} &13.5M &16$\times$3$\times$1  &122 &55.6 &82.1 \\
            \textbf{PosMLP-Video-B} & &19.0M &16$\times$3$\times$1  &177 &58.2 &\textbf{84.6} \\
            \textbf{PosMLP-Video-L} & &35.4M &16$\times$3$\times$1  &338 &\textbf{59.0} &84.3 \\
            % PosMLP-Video-L & & &16$\times$3$\times$1  &--- &--- &--- \\
            % \Xcline{1-7}{0.8pt}
            \bottomrule
		\end{tabular}
  }
		
		% \vspace{-0.2cm}
\end{table*}

{\bf  PoTGU+PoSGU variants.} In Figure \ref{model_posblockversion}, we present four PoTGU+PoSGU block versions (V1-4). These versions incorporate various feature operations such as channel splitting (V2), channel concatenation (V2, V3), and elementwise addition (V4). By adjusting the value of $r_e$, the model size can be conveniently controlled. The experimental results, as presented in Table \ref{tab:posblockversion}, suggest that a larger model size (i.e., more parameters) generally leads to better recognition performance. Interestingly, the use of channel splitting and concatenation (V2) does not improve the top-1 accuracy, despite having similar model sizes when compared to V1. Conversely, although elementwise addition (V4) reduced the channel length significantly, it also led to a degradation in recognition accuracy, as observed in V3 and V4. To further compare V1 and V2, we also implement their base models as shown in the last two rows of Table \ref{tab:posblockversion} and find that V1 also obtains better accuracies than V2.

\begin{table*}[ht]
\caption{Comparison of performance on Something-Something V2 dataset.}
		\label{tab:res_somev2}
		\centering
		% \scriptsize
        % \footnotesize
		% \small
% 		\vspace{-0.2cm}
        
\resizebox{1.9\columnwidth}{!}{
		\begin{tabular}{l|c|c|c|c|cc}
			% \hline
   \toprule
% 			& & & & && \multicolumn{2}{c}{V1} && \multicolumn{2}{c}{V2}\\
%         \cline{7-9}\cline{10-11}
			\textbf{Method} &\textbf{Pretrain} &\textbf{Params} &\textbf{Fr.$\times$Cr.$\times$Cl.} &\textbf{GFLOPs} &\textbf{Top-1} &\textbf{Top-5} \\
            \Xcline{1-7}{0.6pt}
            \multicolumn{7}{c}{\textbf{\textit{CNNs}}} \\
            \hline
            % TIN \cite{shao2020temporal} & &24.6M  &16$\times$1$\times$1 &67.0 &60.1 &86.4 \\
            TSM \cite{lin2019tsm} &\multirow{14}*{IN-1K} &23.9M  &16$\times$1$\times$2 &131.6  &63.1 &88.2  \\
            SlowFast \cite{feichtenhofer2019slowfast} & &53.3M  &40$\times$3$\times$2 &636  &63.1 &87.6 \\
            SmallBig \cite{li2020smallbignet} & &---  &16$\times$3$\times$2 &--- &63.8 &88.9  \\
            STM \cite{jiang2019stm} & &24.0M  &16$\times$3$\times$10 &999 &64.2 &89.8  \\
            AIA(TSM)~\cite{hao2022attention} & &23.9M  &16$\times$3$\times$2 &397.8 &64.3 &88.9  \\
            TANet \cite{liu2021tam} & &25.6M &16$\times$3$\times$2 &--- &64.6 &89.5 \\
            TEINet \cite{liu2020teinet} & &30.4M  &16$\times$1$\times$10 &990 &64.7 &---  \\
            MSNet \cite{kwon2020motionsqueeze} & &24.6M &16$\times$1$\times$1 &67  &64.7 &89.4 \\
            TEA \cite{li2020tea}  & &---  &16$\times$3$\times$10 &2100 &65.1 &89.9  \\
            TDN \cite{wang2021tdn} & &26.1M  &16$\times$1$\times$1 &72.0 &65.3 &89.5  \\
            SDA-TSM~\cite{tan2021selective} & &25.8M  &16$\times$3$\times$2 &406.8 &65.4 &90.0  \\
            CT-NET~\cite{li2021ct}    & &---  &16$\times$3$\times$2 &447 &65.9 &\textbf{90.1}  \\
            GC-TDN \cite{hao2022group} & &27.4M  &16$\times$1$\times$1 &73.4 &65.9 &90.0  \\
            H$^{2}$CN~\cite{tan2022hierarchical} & &24.1M  &16$\times$1$\times$1 &67.6 &\textbf{66.4} &\textbf{90.1}  \\
            \Xcline{1-7}{0.6pt}
            \multicolumn{7}{c}{\textbf{\textit{Transformers}}} \\
            \hline
            TimeSformer-HR~\cite{bertasius2021space} &IN-21K &121.4M  &16$\times$3$\times$1 &5109  &62.5 &--- \\
            ViViT-L/16$\times$2~\cite{arnab2021vivit} &IN-21K &352.1M &16$\times$3$\times$4 &11892  &65.4 &89.8 \\
            DVT~\cite{wang2022deformable} &IN-1K &73.9M &16$\times$3$\times$1 &385  &66.7 &90.8 \\
            % ILA-ViT-B/16~\cite{tu2023implicit} &CLIP-400M &---  &16$\times$4$\times$3 &5256  &66.8 &90.3 \\
            X-ViT~\cite{bulat2021space} &K600 &92.0M &16$\times$3$\times$1 &850  &67.2 &90.8 \\
            MM-ViT~\cite{chen2022mm} &IN-21K &158.1M &16$\times$3$\times$1 &4530  &67.4 &90.6 \\
            MTV-B~\cite{yan2022multiview} &IN-21K &310M  &32$\times$3$\times$1 &963  &67.6 &90.1 \\ %diving 48
            MViT-B~\cite{fan2021multiscale} &K400 &36.6M  &64$\times$3$\times$1 &1365  &67.7 &90.9 \\
            RViT-XL,64$\times$3~\cite{yang2022recurring} &K400 &107.7M &64$\times$3$\times$3 &3990  &67.9 &91.2 \\
            ORViT MF~\cite{herzig2022object}  &IN-21K+K400 &148M  &16$\times$3$\times$1  &405 &67.9 &90.5 \\
            MFormer-L~\cite{patrick2021keeping} &IN-21K+K400 &---  &32$\times$3$\times$1 &3555  &68.1 &91.2 \\
            MViTv2-S,16$\times$4~\cite{li2022mvitv2} &K400 &34.4M  &16$\times$3$\times$1 &194  &68.2 &91.4 \\
            Swin-B~\cite{liu2022video} &K400 &88.8M  &16$\times$3$\times$1 &963  &69.6 &92.7 \\
            VideoMAE-B~\cite{tong2022videomae} &K400 &87.0M &16$\times$2$\times$3 &1080  &69.7 &92.3 \\
            UniFormer-B~\cite{li2023uniformer} &K400 &$\sim$50M  &16$\times$3$\times$1 &290  &\textbf{70.4} &\textbf{92.8} \\
            % \midrule[0.5pt]
            \Xcline{1-7}{0.7pt}
            \multicolumn{7}{c}{\textbf{\textit{MLPs}}} \\
            \hline
            MLP-3D-S \cite{qiu2022mlp} &\multirow{6}*{IN-1K} &74.1M &64$\times$3$\times$1  &324 &67.2 &91.3 \\
            MLP-3D-M \cite{qiu2022mlp} & &88.3M &64$\times$3$\times$1  &549 &68.0 &91.7 \\
            MLP-3D-L \cite{qiu2022mlp} & &149.4M &64$\times$3$\times$1  &1008 &68.5 &92.0 \\
            MorphMLP-S \cite{zhang2022morphmlp} & &46.9M &16$\times$3$\times$1 &201  &67.1  &90.9 \\
            MorphMLP-B \cite{zhang2022morphmlp} & &67.6M &16$\times$3$\times$1 &294  &67.6  &91.3 \\
            MorphMLP-B \cite{zhang2022morphmlp} & &68.5M &32$\times$3$\times$1 &591  &70.1  &\textbf{92.8} \\
            \Xcline{1-7}{0.6pt}
            \textbf{PosMLP-Video-S} &\multirow{3}*{IN-1K} &13.5M &16$\times$3$\times$1  &122 &68.1 &91.3 \\
            \textbf{PosMLP-Video-B} & &19.0M &16$\times$3$\times$1  &177 &70.1 &92.5 \\
            \textbf{PosMLP-Video-L} & &35.4M &16$\times$3$\times$1  &338 &\textbf{70.3} &92.3 \\
            % \textbf{PosMLP-Video-L} & &21.6M &16$\times$3$\times$1  &203 &\textbf{70.5} &92.5 \\
            % \Xcline{1-7}{0.8pt}
            \bottomrule
		\end{tabular}
  }
		% \vspace{-0.3cm}
		% \vspace{-0.4cm}
\end{table*}

The conducted ablation studies have thoroughly addressed the design questions (DQ1-2). The results successfully demonstrate the following insights: (1) Extending the LRPE-based spatial PosMLP for temporal modeling has proven to be effective in capturing temporal information. We validate the necessity of a larger temporal receptive field for robust temporal modeling (DQ1). (2) Through meticulous exploration of various spatio-temporal factorization methods, we have identified more efficient and effective approaches for combining spatial and temporal information (DQ2). In summary, these ablation studies have provided valuable insights into optimizing our model's performance across different dimensions.

 % (3) Implementing a channel grouping strategy has enabled the creation of multiple relative position bias dictionaries, enhancing the diversity of token relationships. This enhancement significantly improves model performance (DQ3).

% zhouds add<-

\begin{figure*}[]
\centering
\includegraphics[width=1.0\textwidth]{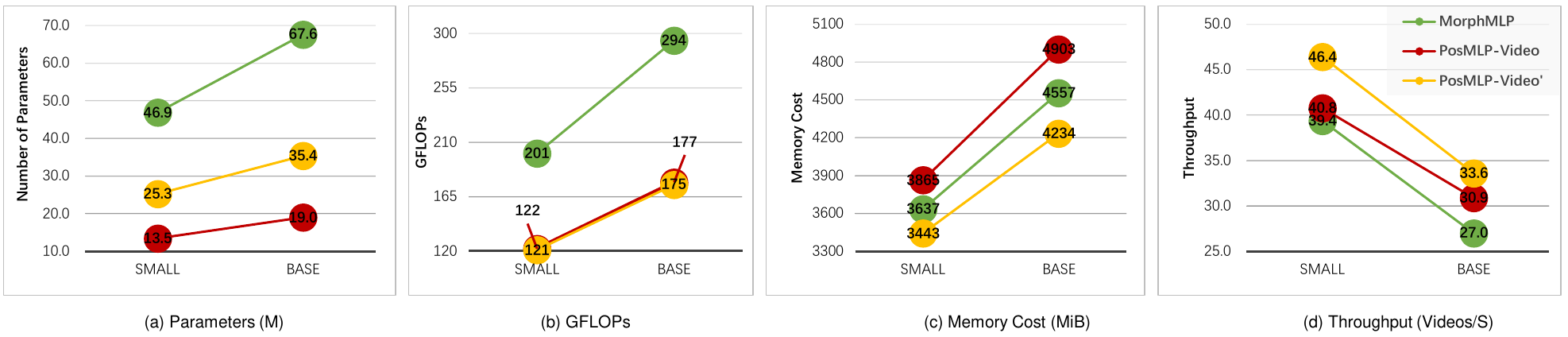}
\caption{Comparison of parameters, FLOPs, memory cost (MiB) and throughput (Videos/S) on SSV1. The input is $16\times 3\times 1$ frames for all methods. PosMLP-Video'-S (Small) and PosMLP-Video'-B (Base) have the same layer numbers, channel settings and time length transformation ($T\rightarrow \frac{T}{2}$) as MorphMLP-S and MorphMLP-B. PosMLP-Video'-S obtains 54.2\% top-1 accuracy compared to 53.9\% of MorphMLP-S.}
\label{comp_curve}
% \vspace{-0.3cm}
\end{figure*}

% \begin{table*}[htb]
% \caption{Comparison of memory cost (MiB) and throughput (videos/S) on SSV1. The input is $16\times 3\times 1$ frames for all methods. PosMLP-Video-S' has the same layer numbers, channel settings $(112, 224, 392, 784)$ and time length transformation ($T\rightarrow \frac{T}{2}$) as MorphMLP-S. PosMLP-Video-S' obtains 54.2\% top-1 accuracy compared to 53.9\% of MorphMLP-S.}
% 		\label{tab:memvs}
% 		\centering
% 		% \scriptsize
%         % \footnotesize
% 		\small
%         % \resizebox{1.05\columnwidth}{!}{
% 		\begin{tabular}{l|c|c|c|c}
% 			% \hline
%    \toprule
% % 			& & & & && \multicolumn{2}{c}{V1} && \multicolumn{2}{c}{V2}\\
% %         \cline{7-9}\cline{10-11}
% 			\textbf{Method} &\textbf{Params.} &\textbf{GFLOPs} &\textbf{Mem. (MiB)} &\textbf{Throughput (Videos/S)} \\
%             \midrule[1pt]
%             % X3D-L \cite{feichtenhofer2020x3d} &--- &20.3M &--$\times$3$\times$10  &5820 &\textbf{80.4} &\textbf{94.6} \\
%             MViTv2-S~\cite{li2022mvitv2}        &34.4M  &320  &2205  &27.88  \\
%             MorphMLP-S \cite{zhang2022morphmlp} &46.9M  &201  &3637  &39.38  \\
%             MorphMLP-B \cite{zhang2022morphmlp} &67.6M  &294  &4547  &27.00  \\
%             \textbf{PosMLP-Video-S} &13.5M  &122 &3865 &40.77 \\
%             \textbf{PosMLP-Video-B} &19.0M  &177 &4903 &30.92 \\
%             \hline
%             \textbf{PosMLP-Video-S'}, $T\rightarrow \frac{T}{2}$, $(112, 224, 392, 784)$ &25.3M  &121 &3443 &46.35 \\
%             \bottomrule
% 		\end{tabular}
%   % }
% \end{table*}

\begin{table*}[htp]
\caption{Comparison of performance on Kinetics-400 dataset.}
		\label{tab:res_kinetic}
		\centering
		% \scriptsize
        % \footnotesize
		% \small
% 		\vspace{-0.2cm}
        \resizebox{1.9\columnwidth}{!}{
		\begin{tabular}{l|c|c|c|c|cc}
			% \hline
   \toprule
% 			& & & & && \multicolumn{2}{c}{V1} && \multicolumn{2}{c}{V2}\\
%         \cline{7-9}\cline{10-11}
			\textbf{Method} &\textbf{Pretrain} &\textbf{Params} &\textbf{Fr.$\times$Cr.$\times$Cl.} &\textbf{GFLOPs} &\textbf{Top-1} &\textbf{Top-5} \\
            % \midrule[1pt]
            \Xcline{1-7}{0.8pt}
            % R(2+1)D \cite{tran2018closer} &ResNet-34  &32$\times$N/A &72.0 &90.0\\
            \multicolumn{7}{c}{\textbf{\textit{CNNs}}} \\
            \hline
            % I3D (InceptionV1) \cite{carreira2017quo} & --- &64 &---  &72.1 &90.3 \\
            TSM \cite{lin2019tsm} &IN-1K &24.3M  &16$\times$1$\times$10 &660 &74.7 &91.4 \\
            NL-I3D \cite{wang2018non} &IN-1K &35.3M  &32$\times$1$\times$10 &2820  &74.9 &91.6 \\
            % S3D-G (InceptionV1) \cite{xie2018rethinking} &--- &64 &71.4G$\times$30 &74.7 &\textbf{93.4} \\
            % I3D-Two-Stream \cite{carreira2017quo} &ResNet-50 &ImageNet &N/A  &75.7 &92.0 \\
            %$\text{ECO}_{En}$ \cite{zolfaghari2018eco} &N/A &70.0 &89.4 \\
            % TEA \cite{li2020tea} & &16$\times$1$\times$10 &70G$\times$30 &76.1 &92.5 \\
            TEINet \cite{liu2020teinet} &IN-1K &30.8M & 16$\times$3$\times$10 &1980 &76.2 &92.5 \\
            TANet-R50 \cite{liu2021tam} &IN-1K &25.6M &16$\times$4$\times$3 &1032 &76.9 &92.9 \\
            % TEA \cite{li2020tea} & ResNet-50 &16$\times$30clip &76.1 &92.5 \\
            SmallBig-R101 \cite{li2020smallbignet} &IN-1K &--- &32$\times$3$\times$4 &5016 &77.4  &93.3 \\
            % SlowFast(4$\times$16) \cite{feichtenhofer2019slowfast} &32.9M & 4+32 &36.1G$\times$30 &75.6 &92.1 \\
            H$^{2}$CN~\cite{tan2022hierarchical}    &IN-1K &24.1M  &16$\times$3$\times$10 &2028 &77.9 &93.3  \\
            TDN-R101 \cite{wang2021tdn} &IN-1K &--- &16$\times$3$\times$10 &3960 &78.5   &93.9 \\
            CT-NET-R101~\cite{li2021ct}    &IN-1K &---  &16$\times$3$\times$4 &1746 &78.8 &93.7  \\
            GC-TDN-R50~\cite{hao2022group} &IN-1K &27.4M &16$\times$3$\times$10 &2202 &78.8  &93.8 \\
            SlowFast101+NL \cite{feichtenhofer2019slowfast} &--- &59.9M &80$\times$3$\times$10 &7020 &79.8 &93.9 \\
            X3D-XXL \cite{feichtenhofer2020x3d} &--- &20.3M &--$\times$3$\times$10  &5820 &\textbf{80.4} &\textbf{94.6} \\
            \Xcline{1-7}{0.7pt}
            \multicolumn{7}{c}{\textbf{\textit{Transformers}}} \\
            \hline
            TokShift~\cite{zhang2021token} &IN-21K &85.9M &16$\times$3$\times$10 &8085 &78.2 &93.8 \\
            SACS-H~\cite{zhang2022long} &IN-21K &40M &32$\times$3$\times$5 &5190 &79.7 &94.1 \\
            Mformer-L~\cite{patrick2021keeping} &IN-21K &---  &32$\times$3$\times$10 &35553  &80.2 &94.8 \\
            ViViT-L/16$\times$2~\cite{arnab2021vivit} &IN-21K &310.8M &16$\times$3$\times$4 &17357  &80.6 &94.7 \\
            MViT-B,16$\times$4~\cite{fan2021multiscale} &--- &36.6M  &16$\times$1$\times$5 &353  &78.4 &93.5 \\
            MViT-B,32$\times$3~\cite{fan2021multiscale} &--- &36.6M  &32$\times$1$\times$5 &850  &80.2 &94.4 \\
            Swin-S~\cite{liu2022video} &IN-1K &49.8M  &32$\times$3$\times$4 &1992  &80.6 &94.5 \\
            Swin-B~\cite{liu2022video} &IN-1K &88.1M  &32$\times$3$\times$4 &3384  &80.6 &94.6 \\
            TimeSformer-L~\cite{bertasius2021space} &IN-21K &121.4M  &96$\times$3$\times$1 &7140  &80.7 &94.7 \\
            X-ViT~\cite{bulat2021space} &IN-21K &92.0M &16$\times$3$\times$2 &1700  &80.7 &94.7 \\
            DVT~\cite{wang2022deformable} &IN-1K &73.9M &16$\times$1$\times$5 &640  &80.8 &95.0 \\
            MViTv2-S,16$\times$4~\cite{li2022mvitv2} &--- &34.5M &16$\times$1$\times$4 &320  &81.0 &94.6 \\
            % ILA-ViT-B/16~\cite{tu2023implicit} &CLIP-400M &---  &8$\times$4$\times$3 &480  &81.3 &95.0 \\
            % ILA-ViT-B/16~\cite{tu2023implicit} &CLIP-400M &---  &16$\times$4$\times$3 &900  &82.4 &95.5 \\
            RViT-XL,32$\times$3$\times$1~\cite{yang2022recurring} &IN-21K &107.7M &32$\times$3$\times$3 &2010  &80.3 &94.4 \\
            RViT-XL,64$\times$3$\times$1~\cite{yang2022recurring} &IN-21K &107.7M &64$\times$3$\times$3 &11900  &81.5 &95.0 \\
            VideoMAE-B~\cite{tong2022videomae} &--- &87.0M &16$\times$5$\times$3 &2700  &81.5 &95.1 \\
            MTV-B~\cite{yan2022multiview} &IN-21K &310M &32$\times$3$\times$4 &4790  &81.8 &95.0 \\
            UniFormer-S~\cite{li2023uniformer} &IN-1K &$\sim$22M &16$\times$1$\times$4 &167  &80.8 &94.7 \\
            UniFormer-B~\cite{li2023uniformer} &IN-1K &$\sim$50M &16$\times$1$\times$4 &389  &\textbf{82.0} &\textbf{95.1} \\
            \Xcline{1-7}{0.7pt}
            \multicolumn{7}{c}{\textbf{\textit{MLPs}}} \\
            \hline
            MorphMLP-S \cite{zhang2022morphmlp} &\multirow{6}*{IN-1K} &47.0M &16$\times$1$\times$4 &268  &78.7  &93.8 \\
            MorphMLP-B \cite{zhang2022morphmlp} & &67.8M &16$\times$1$\times$4 &392  &79.5  &94.4 \\
            MorphMLP-B \cite{zhang2022morphmlp} & &68.5M &32$\times$1$\times$4 &788  &80.8  &94.9 \\
            MLP-3D-S \cite{qiu2022mlp} & &68.5M &64$\times$3$\times$4  &1224 &80.2 &93.8 \\
            MLP-3D-M \cite{qiu2022mlp} & &80.5M &64$\times$3$\times$4  &2040 &81.0 &94.9 \\
            MLP-3D-L \cite{qiu2022mlp} & &135.6M &64$\times$3$\times$4  &3696 &81.4 &95.2 \\
            % MorphMLP-S \cite{zhang2022morphmlp} & & &16$\times$3$\times$1 &201  &53.9  &81.3 \\
            \Xcline{1-7}{0.6pt}
            \textbf{PosMLP-Video-S} &\multirow{6}*{IN-1K} &13.6M &16$\times$1$\times$4  &162 &78.5 &93.9 \\
            \textbf{PosMLP-Video-B} & &19.1M &16$\times$1$\times$4  &236 &80.3 &94.6 \\
            % PosMLP-Video-B & &19.1M &16$\times$3$\times$4  &708 &80.4 &94.8 \\
            % PosMLP-Video-L & &21.7M &16$\times$1$\times$4  &271 &80.8 &94.9 \\ %67.7*4
            % PosMLP-Video-L & &21.7M &16$\times$3$\times$4  &812 &81.0 &95.0 \\ %67.7*4
            \textbf{PosMLP-Video-L} & &35.4M &16$\times$1$\times$4  &450 &81.2 &94.7 \\ %112.5*12
            \textbf{PosMLP-Video-L} & &35.4M &16$\times$3$\times$4  &1350 &81.6 &94.9 \\ %112.5*12
            \textbf{PosMLP-Video-L} & &35.5M &24$\times$1$\times$4  &679 &81.7 &95.2 \\ %112.5*12
            \textbf{PosMLP-Video-L} & &35.5M &24$\times$3$\times$4  &2037 &\textbf{82.1} &\textbf{95.3} \\ %112.5*12
            % \Xcline{1-7}{0.8pt}
            \bottomrule
		\end{tabular}
  }
\end{table*}

\subsection{Comparison with State-of-the-Art}
We compare PosMLP-Video networks with various state-of-the-art networks, including video CNNs, Transformers and MLPs, on many video recognition tasks. All the competing methods adopt RGB frames as input and are pre-trained on ImageNet1K (IN-1K), ImageNet21K (IN-21K), Kinetics-400 (K400), Kinetics-600 (K600) or None. 
% We report the Top-1/5 accuracy, FLOPs and model parameters for a comprehensive comparison.

{\bf Something-Something V1\&V2.} The two datasets share the same human-performing action categories and only differ in scale. Their video actions focus on more temporal relationships, for example, ``Putting something ...'', ``Lifting something ...'' and ``Pretending to ...''. Tables \ref{tab:res_somev1} and \ref{tab:res_somev2} show the performance comparison on V1 and V2, respectively. On the smaller SSV1, our PosMLP-Video-S achieves a higher Top-1 accuracy of 55.6\% compared to all the competing video CNNs (47.6\%-55.0\%). In comparison with other video MLPs, such as MorphMLP and MLP-3D, PosMLP-Video variants (S, B, L) consistently outperform them and spend much lower computations. For example, PosMLP-Video-L achieves the highest Top-1 accuracy of 59.0\% with only 35.4M parameters/338G FLOPs, which significantly surpasses MLP-3D-L's 56.5\% with 149.4M parameters/1008G FLOPs and MorphMLP-B's 57.4\% with 68.5M parameters/591G FLOPs. Particularly, compared to the other MLP-based MorphMLP, the notable performance improvements further demonstrate the superiority of the proposed positional regimes for spatio-temporal relation modeling.

\begin{table*}[htb]
\caption{Comparison of performance on Diving48 dataset.}
		\label{tab:res_div48}
		\centering
		% \scriptsize
        \footnotesize
		% \small
        % \resizebox{1.05\columnwidth}{!}{
        
		\begin{tabular}{l|c|c}
			% \hline
        \toprule
% 			& & & & && \multicolumn{2}{c}{V1} && \multicolumn{2}{c}{V2}\\
%         \cline{7-9}\cline{10-11}
			\textbf{Method} &\textbf{Fr.$\times$Cr.$\times$Cl.} &\textbf{Top-1} \\
            \Xcline{1-3}{0.6pt}
            \multicolumn{3}{c}{\textbf{\textit{CNNs}}} \\
            \hline
            SlowFast-R101,16x8 \cite{feichtenhofer2019slowfast} from \cite{bertasius2021space} &(64+16)$\times$3$\times$1  &77.6 \\
            AIA(TSM)~\cite{hao2022attention} &8$\times$1$\times$1 &79.4  \\
            TQN \cite{zhang2021temporal} &all frames  &81.8 \\
            GC-TDN \cite{hao2022group} &16$\times$1$\times$1 &87.6  \\
            TFCNet \cite{zhang2022tfcnet} &32$\times$3$\times$1 &88.3  \\
            \Xcline{1-3}{0.6pt}
            \multicolumn{3}{c}{\textbf{\textit{Transformers}}} \\
            \hline
            TimeSformer-L~\cite{bertasius2021space} &96$\times$3$\times$1 &81.0 \\
            VIMPAC~\cite{tan2021vimpac} &--$\times$3$\times$10 &85.5 \\
            % BEVT~\cite{}  &    &86.7 \\
            ORViT TimeSformer \cite{herzig2022object} &32$\times$3$\times$1  &88.0 \\
            % TFCNet~\cite{zhang2022tfcnet} &32$\times$3$\times$1   &88.3 \\
            % MorphMLP-S \cite{zhang2022morphmlp} & & &16$\times$3$\times$1 &201  &53.9  &81.3 \\
            \Xcline{1-3}{0.6pt}
            \multicolumn{3}{c}{\textbf{\textit{MLPs}}} \\
            \hline
            \textbf{PosMLP-Video-L} &16$\times$1$\times$1   &\textbf{88.9} \\
            % \Xcline{1-7}{0.8pt}
            \bottomrule
		\end{tabular}
  % }
		% \vspace{-0.5cm}
\end{table*}

\begin{table*}[htb]
\caption{Comparison of performance on EGTEA Gaze+ dataset.}
		\label{tab:res_egaze}
		\centering
		% \scriptsize
        \footnotesize
		% \small
        % \resizebox{1.05\columnwidth}{!}{
		\begin{tabular}{l|c|c}
			% \hline
        \toprule
% 			& & & & && \multicolumn{2}{c}{V1} && \multicolumn{2}{c}{V2}\\
%         \cline{7-9}\cline{10-11}
			\textbf{Method} &\textbf{Fr.$\times$Cr.$\times$Cl.} &\textbf{Top-1} \\
            \Xcline{1-3}{0.6pt}
            \multicolumn{3}{c}{\textbf{\textit{CNNs}}} \\
            \hline
            % TSM~\cite{} &8 &63.5 \\
            % BEVT~\cite{}  &    &86.7 \\
            SAP~\cite{wang2020symbiotic} &64$\times$1$\times$1  &64.1 \\
            GST-R50~\cite{luo2019grouped} &8$\times$1$\times$1   &64.4 \\
            AIA(TSM)~\cite{hao2022attention} &8$\times$1$\times$1 &64.7 \\
            GC-TSM~\cite{hao2022group} &8$\times$1$\times$1   &66.5 \\
            \Xcline{1-3}{0.6pt}
            \multicolumn{3}{c}{\textbf{\textit{Transformers}}} \\
            \hline
            ViT (Video) \cite{dosovitskiy2020image} from \cite{zhang2021token} &8$\times$1$\times$1 &62.6 \\
            TokShift (HR)~\cite{zhang2021token} &8$\times$1$\times$1 &65.8 \\
            LAPS (H)~\cite{zhang2021token} &32$\times$1$\times$1 &66.1 \\
            % BEVT~\cite{}  &    &86.7 \\
            % MorphMLP-S \cite{zhang2022morphmlp} & & &16$\times$3$\times$1 &201  &53.9  &81.3 \\
            \Xcline{1-3}{0.6pt}
            \multicolumn{3}{c}{\textbf{\textit{MLPs}}} \\
            \hline
            \textbf{PosMLP-Video-L} &16$\times$1$\times$1   &\textbf{72.5} \\
            % \Xcline{1-7}{0.8pt}
            \bottomrule
		\end{tabular}
  % }
		% \vspace{-0.5cm}
\end{table*}

On the larger SSV2, PosMLP-Video variants consistently outperform video CNNs, most Transformers (except UniFormer-B) and MLPs. In particular, PosMLP-Video-L pre-trained on IN-1K  achieves the top-1 accuracy of 70.3\% with $16\times3\times1$ frames input, which even outstrips the video Transformers such as X-ViT, RViT-XL, MFormer-L, MViTv2-S and Swin-B that are pra-trained on the larger-scale datasets (e.g., IN-21K, K400 and K600) and use more frames (e.g., $32\times3\times1$ and $64\times3\times1/3$) as input. More importantly, PosMLP-Video-L has only 35.4M parameters, which is 40\% of Swin-B and 70\% of UniFormer-B. The FLOPs of PosMLP-Video-L is 338G, which is only 35\% of Swin-B. Moreover, compared to the video MLPs, i.e., MLP-3D and MorphMLP, our PosMLP-Video, regardless of network versions, consistently outperforms them with large performance improvements (0.2\%-2.5\%) while requiring much less computational costs (about 18\%-51\% parameters and 33\%-60\% FLOPs).

We also undertake a thorough comparison of model size and computational costs between the representative MLP-like model MorphMLP and our PosMLP-Video. As demonstrated in Figure \ref{comp_curve}, despite the significantly reduced parameters and FLOPs, our standard PosMLP-Video models demonstrate relatively higher memory requirements compared to MorphMLP models. We attribute this to two main reasons. The first reason is the bottleneck structure, which determines that the reduction of memory cost may not be substantial. The second reason is that MorphMLP reduces the time length (number of input frames) from $T$ to $\frac{T}{2}$ after patch embedding, meaning that only half of the input frames are processed in the main network of the model. To verify this, we also adopt the same time-reduction strategy, resulting in the model variants PosMLP-Video’-S and PosMLP-Video’-B. It can be observed that PosMLP-Video’-S and PosMLP-Video’-B consistently achieve fewer parameters, lower FLOPs, lower memory cost, and higher throughput than MorphMLP-S and MorphMLP-B. And PosMLP-Video’-S achieves a higher top-1 accuracy of 54.2\% compared to MorphMLP-S's 53.9\%.

% of memory cost and throughput with other models on SSV1 dataset during inference. Table \ref{tab:memvs} shows the results. Our PosMLP-Video models, with default settings, exhibit relatively higher memory requirements compared to MorphMLP models but achieve superior throughput speeds. Additionally, we provide PosMLP-Video-S' ', configured with identical channel numbers and a time length transformation of $T\rightarrow \frac{T}{2}$ as MorphMLP-S. Notably, PosMLP-Video-S' outperforms MorphMLP-S across all metrics (parameters, FLOPs, memory usage, and throughput), boasting a higher top-1 accuracy of 54.2\% compared to MorphMLP-S's 53.9\%.

{\bf Kinetics-400.} K400 is a large-scale video recognition dataset, whose video categories depend not so much on temporal relations. We list the performance comparison with the SOTA methods in Table \ref{tab:res_kinetic}. Similar to observations as on SSV1 and SSV2, PosMLP-Video models perform consistently better than the competing video CNNs such as GC-TDN, SlowFast101+NL and X3D-XXL. By comparing with Transformer-based models, our PosMLP-Video-L pre-trained on ImageNet-1K achieves the best top-1 accuracy of 82.1\% with $24\times3\times4$ frames. Compared to MTV-B which obtains 81.8\% top-1 accuracy pre-trained on ImageNet-21K and has 310M parameters and 4790 GFLOPs, PosMLP-Video-L only requires 35.5M parameters and 2037 GFLOPs. In other words, the model size of our PosMLP-Video-L is as small as 11\% of MTV-B's, and the computational cost is only its 43\%.  UniFormer-B, which obtains the second best 82.0\% top-1 accuracy with $16\times 1\times 4$ frames, requires only 389G FLOPs. However, its model parameter is about 50M, larger than our PosMLP-Video-L's 35.4M. Also, compared to MorphMLP and MLP-3D, the proposed PosMLP-Video obtains higher recognition performance and requires less computational burden.

{\bf Diving48.} It contains unambiguous dive sequences. Dives entail several stages and necessitate long-range temporal modeling. Table \ref{tab:res_div48} provides a performance comparison with other competing methods. The results show that our proposed PosMLP-Video-L model with 16 frames attains the highest Top-1 accuracy of 88.9\% among the competing methods, which include both video CNNs and Transformers. Particularly, the proposed model outperforms the CNN-based TFCNet's accuracy of 88.3\% by 0.6\% and the Transformer-based ORViT TimeSformer's accuracy of 88.0\% by 0.9\%.

{\bf  EGTEA Gaze+.} This dataset consists of videos showing cooking activities that involve intricate spatio-temporal hand-object and object-object interactions. Table \ref{tab:res_egaze} presents a comparison of the results obtained from different methods. Our PosMLP-Video-L achieves the highest Top-1 accuracy of 72.5\%. This remarkable result significantly surpasses the previous methods by a considerable margin (6.0\%-9.9\%), which provides compelling evidence for the superior ability of our model in spatio-temporal modeling.

\subsection{Visualization}
\begin{figure}[]
\centering
\includegraphics[width=0.48\textwidth]{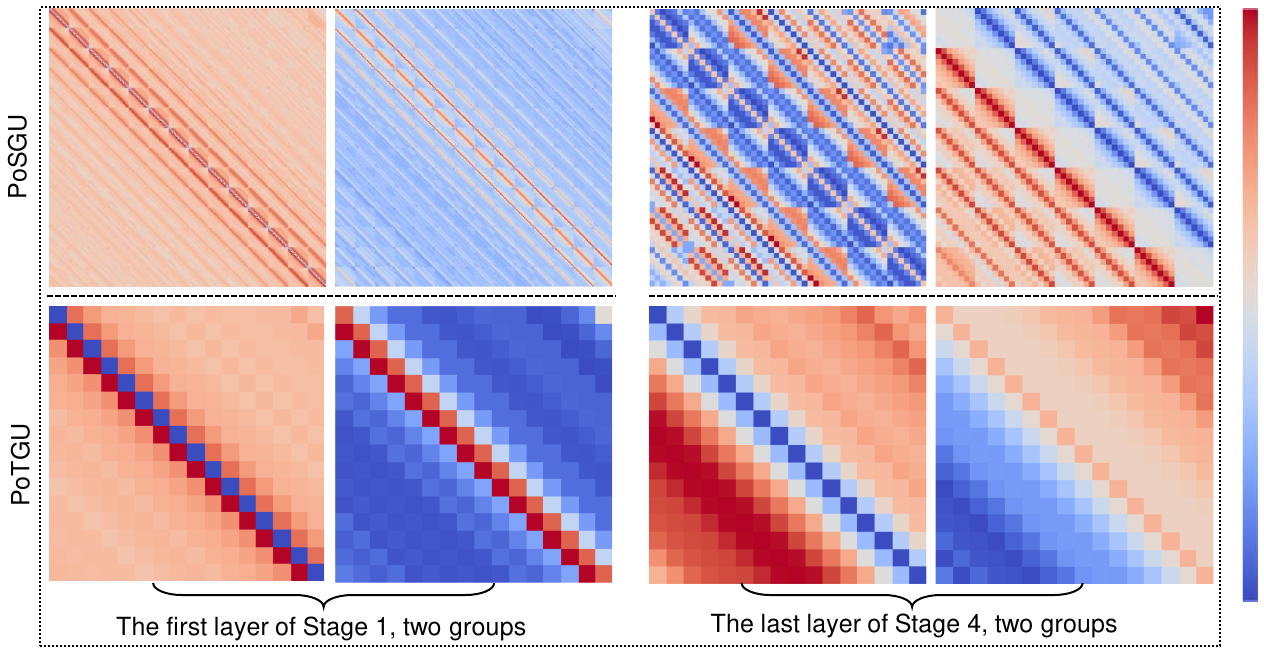}
% \vspace{-0.2cm}
\caption{Visualization of the pairwise token relation matrix learned by PoSGU and PoTGU of PosMLP-Video-S on SSV2 dataset.}
\label{visual}
\vspace{-0.3cm}
\end{figure}

Figure \ref{visual} displays the pairwise token relation matrix learned by PoSGU and PoTGU of PosMLP-Video-S on the SSV2 dataset. We select the first and last layers of PosMLP-Video-S and showcase the learned spatial and temporal token-to-token relations of the two channel groups. Firstly, it can be found that different layers and groups learn various relationship types, indicating that the channel grouping mechanism successfully enriches the relative position relationship types. Secondly, by comparing the relation pattern differences between the first and last layers, we observe that high relation scores mainly exist in local spatial and temporal neighborhoods in the first layer, while they spread all over the region in the last layer. This can be attributed to the fact that there is no cross-token interaction in the first layer, whereas both spatial and temporal tokens have been fully fused in the last layer.

\begin{figure*}
\centering
\subfigure[Moving something closer to something]{
\label{examp1}
\includegraphics[width=1.0\textwidth]{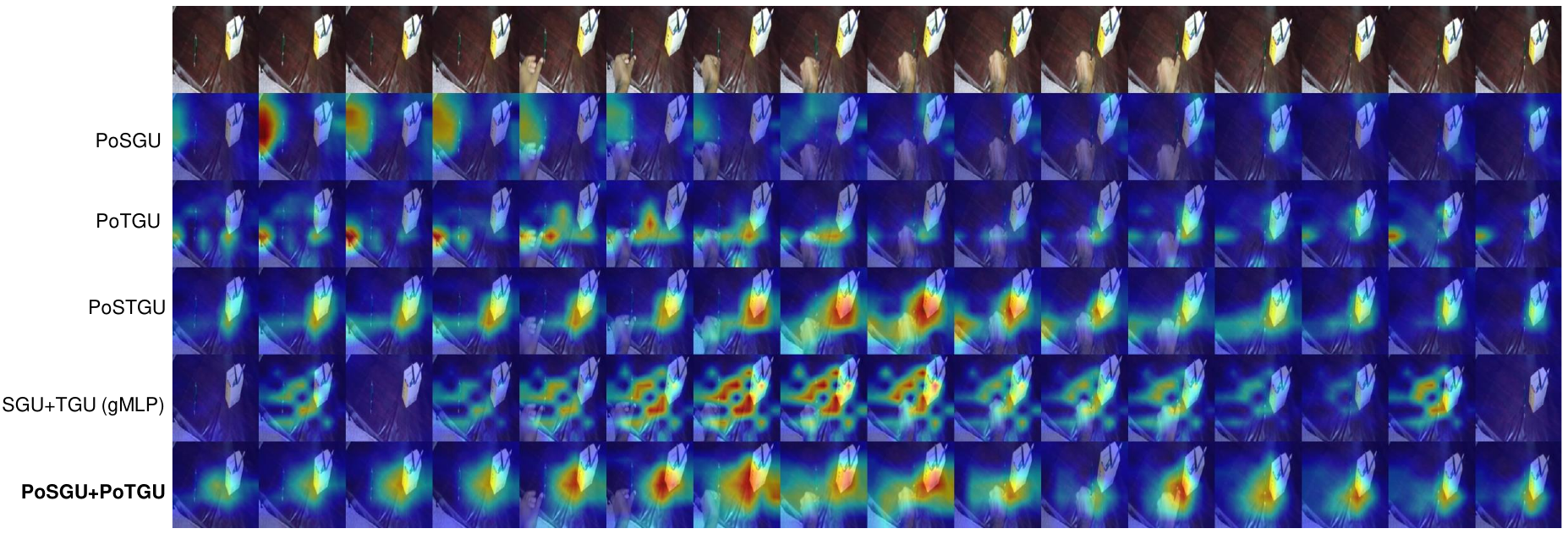}
}
\subfigure[Pushing something from right to left]{
\label{examp2}
\includegraphics[width=1.0\textwidth]{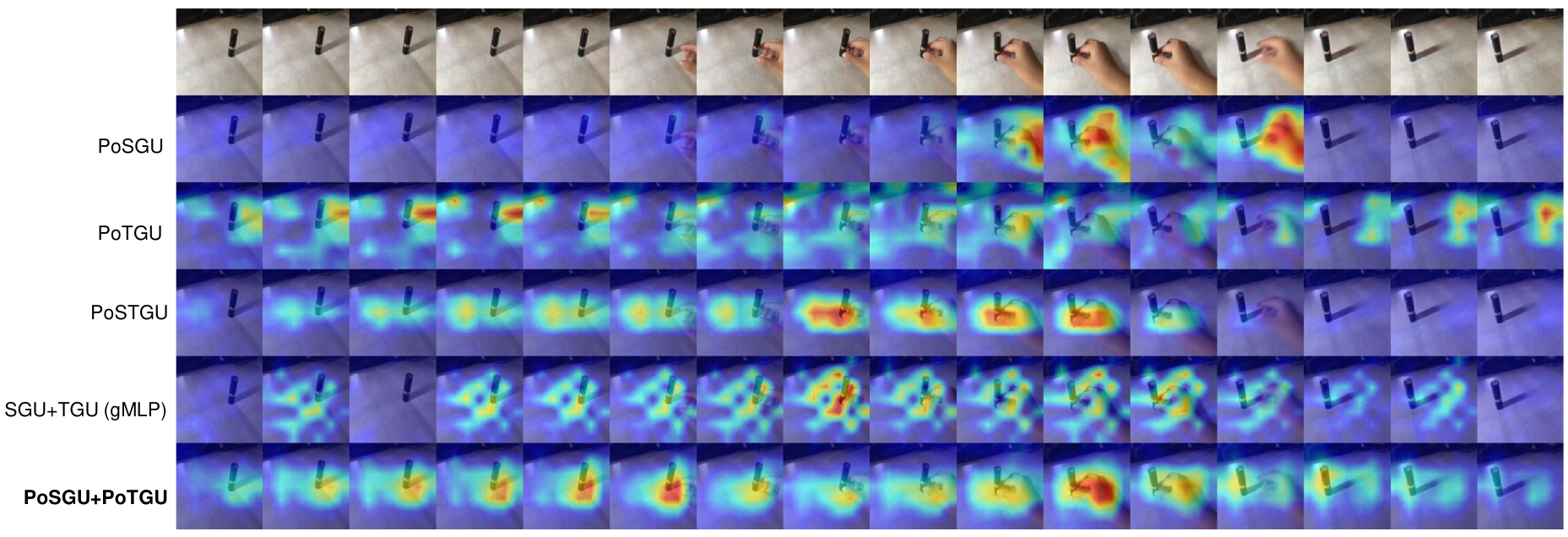}
}
\subfigure[Turning something upside down]{
\label{examp3}
\includegraphics[width=1.0\textwidth]{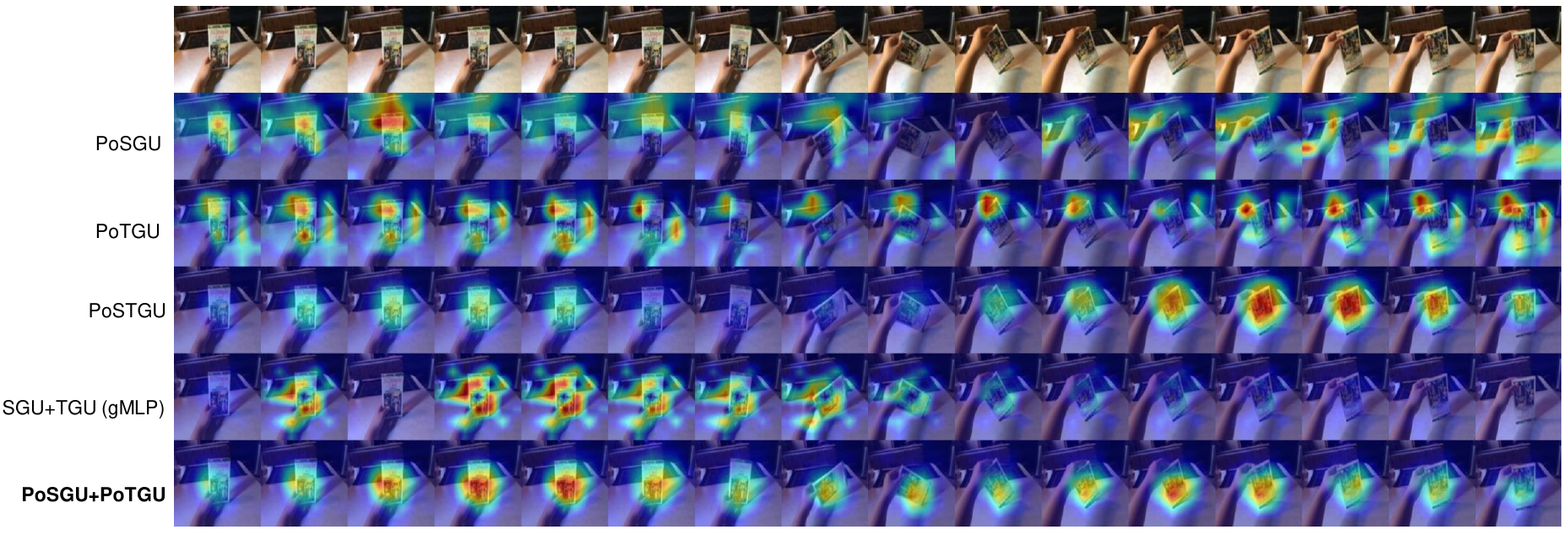}
}
\caption{Visualization examples of class activation maps on SSV1 dataset. The first row shows the original 16 frames.}
\label{visualExamples}
% \vspace{-0.4cm}
\end{figure*}

In Figure \ref{visualExamples}, we present the heatmaps obtained from visualizing the class activation maps of different video PosMLP blocks using the Grad-CAM \cite{selvaraju2017grad} technique. We select action categories with varying moving directions, such as ``\textit{Moving something closer to something}'', ``\textit{Pushing something from right to left}'', and ``\textit{Turning something upside down}'', which involve short-term and long-term temporal interactions between objects. Our observations from the figure reveal that the paralleled PoTGU+PoSGU focuses more on the crucial areas of video frames when compared to both the single pos blocks (PoTGU/PoSGU/PoSTGU) and the similar combined SGU+TGU (gMLP).

% zhouds add<-

\begin{table}[]
\caption{Comparison of action detection performance on AVA v2.2 dataset. All methods are pretrained on K400 dataset. The result of MorphMLP-S is obtained by ourselves.}
		\label{tab:res_ava}
		\centering
        % \tiny
		% \scriptsize
        \footnotesize
		% \small
        % \resizebox{1.05\columnwidth}{!}{
		\begin{tabular}{l|cc|c}
			% \hline
   \toprule
		\textbf{Method}  &\textbf{Params}  &\textbf{GFLOPs} &\textbf{mAP}  \\
   \midrule[1pt]
            SlowFast101 \cite{feichtenhofer2019slowfast}  &53.0M &137.7 &23.8 \\
            MViTv2-S~\cite{li2022mvitv2}  &34.3M &64.5 &26.8 \\
            MorphMLP-S~\cite{zhang2022morphmlp} &46.8M &67.0   &24.7 \\
            \textbf{PosMLP-Video-S}  &13.5M &40.0   &25.2 \\
            \bottomrule
		\end{tabular}
  % }
		% \vspace{-0.2cm}
\end{table}

\begin{table*}[]
\caption{Comparison of MER performance on SMIC, SAMM, CASME II and CASME3 datasets. }
		\label{tab:res_mer}
		\centering
		% \scriptsize
        % \footnotesize
		\small
        % \resizebox{1.05\columnwidth}{!}{
		\begin{tabular}{l|cc|cc|cc|cc}
			% \hline
   \toprule
		\multirow{2}*{\textbf{Method}}   &\multicolumn{2}{c}{\textbf{SMIC}} &\multicolumn{2}{c}{\textbf{SAMM}} &\multicolumn{2}{c}{\textbf{CASME II}} &\multicolumn{2}{c}{\textbf{CASME3}}\\
           &UF1  &UAR  &UF1  &UAR  &UF1  &UAR  &UF1  &UAR \\
   \midrule[1pt]
            MMNet~\cite{li2022mmnet} &44.1 &43.8 &32.6 &34.2 &71.9 &89.9 &--- &--- \\
            STSTNet~\cite{liong2019shallow}  &68.0 &70.1 &65.9 &68.1 &83.8 &86.9 &--- &--- \\
            MERSiamC3D~\cite{zhao2021two}  &73.6 &76.0 &74.8 &72.8 &89.2 &88.7 &--- &--- \\
            RGB-D~\cite{li2022cas} &--- &--- &--- &--- &--- &---  &17.7 &18.3 \\
            Micro-BERT~\cite{nguyen2023micron}   &\textbf{85.3} &\textbf{83.8} &\textbf{83.9} &\textbf{84.8} &90.3 &89.1 &32.6 &32.5\\
            \hline
            MorphMLP-S~\cite{zhang2022morphmlp}  &73.8 &72.9  &74.3 &74.1  &87.4 &86.7 &30.4  &29.8  \\
            \textbf{PosMLP-Video-S}  &82.8 &76.4 &82.1 &81.4 &\textbf{92.1} &\textbf{90.6} &\textbf{33.7}  &\textbf{33.2} \\
            \bottomrule
		\end{tabular}
  % }
		% \vspace{-0.2cm}
\end{table*}

\section{Experiment on Action Detection}

{\bf Dataset and Task.} The AVA dataset~\cite{gu2018ava} provides an action detection task, where the goal is to locate spatio-temporal human actions within videos. To evaluate the performance of different methods, we utilize the AVA v2.2 version and follow the standard evaluation protocol used by SlowFast~\cite{fan2020pyslowfast}. 

{\bf Pipeline and Evaluation.} The action detection pipeline adopts a framework similar to MViTv2 but replaces its backbone with our PosMLP-Video model. We also leverage Faster R-CNN~\cite{ren2015faster} for the task of video action detection.  Here is a breakdown of the workflow: (1) In the first step, we extract region-of-interest (RoI) features~\cite{girshick2015fast} using frame-wise RoIAlign~\cite{he2017mask} on the spatio-temporal maps generated by the last layer of the PosMLP-Video model; and (2) Next, we apply max-pooling to the RoI features and feed the processed features to a per-class action predictor with sigmoid activation. For the evaluation, we adopt the mean Average Precision (mAP) metric to report the performance, which is computed across 60 distinct action classes.

{\bf Training\&Inference.} The training process closely aligns with \cite{li2022mvitv2,fan2020pyslowfast}. We select proposals that have overlaps with ground-truth bouding-boxes of IoU larger than 0.9 for training use. The training configurations are set as follows: a batch-size of 8 clips per GPU, 30 epochs with linear warm-up for the first 5 epochs, an initial learning rate of 0.6 with decayed with $10^{-8}$, a drop-path rate of 0.4, and the SGD optimizer. The inference is performed on a single clip with 16 center-cropped frames.

{\bf Result Comparison.} Table \ref{tab:res_ava} presents the action detection results of SlowFast101, MViTv2-S, MorphMLP-S and our PosMLP-Video-S. We observe that our PosMLP-Video-S achieves a higher validation mAP of 25.6 compared to SlowFast101 (23.8 mAP) and MorphMLP-S (24.7 mAP), both of which are CNN-based and MLP-based models, respectively. Although our PosMLP-Video-S falls slightly behind MViTv2-S (26.8 mAP), it is important to note that our model is significantly more lightweight, with only 13.5M parameters and 40.0 GFLOPs, in contrast to the other three models. Hence, considering this factor, our PosMLP-Video-S delivers comparable performance to MViTv2-S in terms of action detection.

\section{Experiment on Micro-Expression Recognition}

{\bf Dataset and Task.} Micro-Expression Recognition (MER) is a specialized task focused on identifying emotions conveyed through short and subtle facial movements observed in video clips. Therefore, this task emphasizes the model's capability to capture micro spatio-temporal variations. Four datasets are commonly used in MER, such as SMIC~\cite{li2013spontaneous}, SAMM~\cite{davison2016samm}, CASME II~\cite{yan2014casme} and CASME3~\cite{li2022cas}. Specifically, SMIC, SAMM, CASME II and CASME3 contain 161, 159, 247 and $\sim$1000 ME videos, respectively. SMIC, SAMM and CASME II have three ME categories: ``Negative'', ``Positive'' and ``Surprise'', while CASME3 features seven distinct emotional expressions: ``Happiness'', ``Anger'', ``Sadness'', ``Surprise'', ``Fear'', ``Disgust'' and ``Others''.

{\bf Pipeline and Evaluation.} The MER pipeline is similar to video action recognition, where each video clip is assigned a specific emotion. In implementation, we uniformly select six frames from a video clip and input them into the video models to predict the emotion category. For evaluation, we use the leave-one-subject-out cross-validation (LOSO) strategy~\cite{li2022mmnet} on the four datasets. Specifically, during each iteration, one subset is randomly designated as the test set while the remaining subsets constitute the training set. This method ensures robust evaluation across different subjects and minimizes bias. To measure performance, we use evaluation metrics such as unweighted F1-scores (UF1) and unweighted average recall (UAR), which are employed in the MER2019 challenge~\cite{khor2019dual}. 

% Moreover, we also use the more efficient $K$-fold cross-validation strategy~\cite{zhao2023dfme} to compare our model with other general video models.

{\bf Training\&Inference.} The training configurations are set as follows: a batch-size of 8 clips per GPU, 150 epochs with a warm-up for the first 5 epochs, a base learning rate of $5e-5$ and a weight decay of 0.05, and the AdamW optimizer. The inference is performed on a single clip with 6 center-cropped frames.

{\bf Result Comparison.} In Table~\ref{tab:res_mer}, we show the MER results of several MER-specific methods, including MMNet, STSTNet, MERSiamC3D, RGB-D and Micro-BERT, as well as the general video mode MorphMLP-S and our PosMLP-Video-S. MMNet introduces a continuous attention block and a position calibration module to recognize MEs on specific frames, such as apex and onset frames. STSTNet, on the other hand, extracts multiple optical flow features from a video clip and employs a shallow 3D CNN for emotion recognition. MERSiamC3D follows a two-stage learning methodology, where generic ME features are first extracted during the prior learning stage, and the model is subsequently adjusted for high-level feature learning during the target learning stage. RGB-D incorporates depth information into AlexNet to boost RGB features. Micro-BERT integrates their proposed diagonal micro-attention mechanism and a patch of interest module into the BERT network for facial Micro-Expression Recognition (MER). Our PosMLP-Video-S model demonstrates superior performance, achieving the highest results on the larger CASME II and CASME3 datasets and the second-best results on the SMIC and SAMM datasets. Compared to the closely related video MLP model MorphMLP-S, our PosMLP-Video-S consistently outperforms it across all metrics and datasets.

\section{Conclusion}
We have presented a novel MLP-like architecture, PosMLP-Video, for efficient and effective video recognition. Our approach leverages the relative position encoding as a key component, leading to improved pairwise token relations modeling. We introduced a family of positional spatial and temporal gating units (PoTGU, PoSGU, and PoSTGU) that are both more parameters- and FLOPs-efficient than self-attention and token-mixing layers. These units are integrated into spatio-temporal factorized network blocks to promote spatial and temporal modeling. To thoroughly evaluate the effectiveness of our method, we conducted experiments on three video-related tasks: general video recognition, action detection, and micro-expression recognition (MER). These tasks provide different perspectives on video understanding. The general video recognition task involves recognizing a wide range of actions in videos, including both coarse- and fine-grained actions. The superior results on various video recognition datasets convincingly demonstrate our PosMLP-Video's ability in identifying different levels of action granularity. The results from our action detection experiments highlight the impressive capability of our model to accurately locate and identify spatio-temporal actions within a video. MER requires sensitivity to subtle facial dynamics. By successfully recognizing micro-expressions, our model demonstrated its proficiency in capturing and interpreting delicate facial movements that convey emotions.  Overall, these comprehensive experiments showcase the efficacy of the proposed PosMLP-Video and its potential for advancing the field of video understanding.

\bmhead{Acknowledgments}
The work was supported by the National Natural Science Foundation of China (No. 62101524). We thank Prof. Xiangnan He for providing thoughtful comments and suggestions on the writing and structural organization of the paper.

%%===========================================================================================%%
%% If you are submitting to one of the Nature Portfolio journals, using the eJP submission   %%
%% system, please include the references within the manuscript file itself. You may do this  %%
%% by copying the reference list from your .bbl file, paste it into the main manuscript .tex %%
%% file, and delete the associated \verb+\bibliography+ commands.                            %%
%%===========================================================================================%%

\bibliography{egbib}% common bib file
%% if required, the content of .bbl file can be included here once bbl is generated
%%\input sn-article.bbl

\end{document}